%% file: reeve18.tex
\documentclass[tablecaption=bottom,wcp]{jmlr}

\usepackage[load-configurations=version-1]{siunitx}

\theorembodyfont{\upshape}
\theoremheaderfont{\scshape}
\theorempostheader{:}
\theoremsep{\newline}

\usepackage{hyperref}
\usepackage{bm}

\usepackage{subfiles}

\usepackage{enumitem}
\usepackage{verbatim}
\newtheorem{assumption}{Assumption}
\newtheorem{defn}{Definition}
\newtheorem{prop}{Proposition}
\newtheorem{myLemma}{Lemma}
\usepackage{mathtools}
\usepackage{dsfont}

\usepackage{thmtools}
\usepackage{thm-restate}
\usepackage{booktabs}
\usepackage{natbib}
\usepackage{enumitem}

\newcommand{\R}{\mathbb{R}}
\newcommand{\I}{\mathcal{I}}

\newcommand{\supp}{\text{supp}}
\newcommand{\X}{\mathcal{X}}

\newcommand{\Prob}{\mathbb{P}}
\newcommand{\N}{\mathbb{N}}
\newcommand{\E}{\mathbb{E}}
\newcommand{\D}{\mathcal{D}}
\newcommand{\one}{\mathds{1}}
\newcommand{\argmin}{\text{argmin}}
\newcommand{\argmax}{\text{argmax}}
\newcommand{\Xdim}{d}
\newcommand{\diam}{\text{diam}}

\jmlrvolume{83}
\jmlryear{2018}
\jmlrsubmitted{October 27, 2017}
\jmlrpublished{April 7, 2018}
\jmlrworkshop{ALT 2018} % W&CP title

\title[$k$-Nearest Neighbour UCB]{The $k$-Nearest Neighbour UCB Algorithm for Multi-Armed \titlebreak Bandits with Covariates}

  \author{\Name{Henry WJ Reeve}\\ \Email{henry.reeve@manchester.ac.uk}\\ \addr{School of Computer Science, \\
       The University of Manchester,\\ Manchester, UK}\\ \linebreak
\Name{Joe Mellor}\\ \Email{joe.mellor@ed.ac.uk}\\ \addr{  Usher Institute of Population Health Sciences and Informatics\\
    The University of Edinburgh,\\ Edinburgh, UK}\\ \linebreak
\Name{Gavin Brown}\\ \Email{gavin.brown@manchester.ac.uk}\\ \addr{School of Computer Science, \\
       The University of Manchester,\\ Manchester, UK}
 }

\editor{Mehryar Mohri \&
Karthik Sridharan}

\begin{document}

\maketitle

\begin{abstract}
\subfile{Sections/abstract}    
\end{abstract}

\RestyleAlgo{boxruled}

\section{Introduction}\label{introsec}
\subfile{Sections/introduction}

\newpage
\section{Bandits on a metric space}\label{problemSetupSec}
\subfile{Sections/problemSetup}

\subfile{Sections/assumptions}

\section{Nearest neighbour algorithms}\label{algorithmDescSec}
\subfile{Sections/nearestNeighbourAlgorithms}

\newpage
\section{Regret analysis}\label{regretAnalysisSec}
\subfile{Sections/analysisForGeneralisedIndexAlgorithm}

\newpage
\section{Experimental results}\label{experimentalSec}
\subfile{Sections/experimentalResults}

\section{Discussion}\label{discussionSec}
\subfile{Sections/discussion}

\newpage
\section*{Acknowledgements}
\subfile{Sections/acknowledgements}

\bibliographystyle{plain}  
\bibliography{mybib}

\newpage
\appendix

\section{Proof of Theorem \ref{knnUCBRegretBound}}\label{proofOfKnnUCBRegretBoundSec}
\subfile{Sections/knnUCBSubgaussianRegretBound}

\section{Proof of Theorem \ref{knnKlUCBRegretBound}}\label{proofOfKnnKlUCBRegretBoundSec}
\subfile{Sections/knnKlUCBRegretBound}

\section{Concentration Inequalities}\label{concentrationInequalitySec}
\subfile{Sections/ConcentrationInequality}

\section{Local Regret Lemmas}\label{localRegretSec}
\subfile{Sections/localisedRegretLemmas}

\section{Nested Partitions Lemma}\label{nestedPartitionSec}

\subfile{Sections/nestedPartitionsLemma}

\section{Experimental Procedure}\label{experimentalDetailsSec}
\subfile{Sections/experimentalDetails}

\section{Dyadic sub-intervals}\label{exampleDiadicSmallMeasureAppendix} 
\subfile{Sections/exampleDiadicMeasureBadlyBehaved}

\section{Manifolds and the minimax lower bound}\label{manifoldsSec}
\subfile{Sections/manifoldsAndTheLowerBound}

\end{document}

%% file: Sections/abstract.tex
In this paper we propose and explore the $k$-Nearest Neighbour UCB algorithm for multi-armed bandits with covariates. We focus on a setting where the covariates are supported on a metric space of low intrinsic dimension, such as a manifold embedded within a high dimensional ambient feature space. The algorithm is conceptually simple and straightforward to implement. The $k$-Nearest Neighbour UCB algorithm does not require prior knowledge of the either the intrinsic dimension of the marginal distribution or the time horizon. We prove a regret bound for the $k$-Nearest Neighbour UCB algorithm which is minimax optimal up to logarithmic factors. In particular, the algorithm automatically takes advantage of both low intrinsic dimensionality of the marginal distribution over the covariates and low noise in the data, expressed as a margin condition. In addition, focusing on the case of bounded rewards, we give corresponding regret bounds for the $k$-Nearest Neighbour KL-UCB algorithm, which is an analogue of the KL-UCB algorithm adapted to the setting of multi-armed bandits with covariates. Finally, we present empirical results which demonstrate the ability of both the $k$-Nearest Neighbour UCB and $k$-Nearest Neighbour KL-UCB to take advantage of situations where the data is supported on an unknown sub-manifold of a high-dimensional feature space.

%% file: Sections/introduction.tex
The multi-armed bandit is a simple model which exemplifies the exploitation-exploration trade-off in reinforcement learning. Solutions to this problem have numerous practical applications from sequential clinical trials to web-page ad placement (\cite{bubeck2012regret}). We focus upon the stochastic setting in which an agent is given access to a collection of unknown reward distributions (arms); the agent sequentially selects a reward distribution to sample from, so as to maximise their cumulative reward. One of the most widely used strategies for stochastic multi-armed bandits is the Upper Confidence Bound (UCB) algorithm, which is based on the principle of optimism in the face of uncertainty (\cite{lai1985asymptotically,agrawal1995sample,auer2002finite}). Garivier and Capp\'{e}'s KL-UCB algorithm utilises tighter upper confidence bounds to provide an algorithm with sharper regret bounds and a superior empirical performance (\cite{garivier2011kl}).

Multi-armed bandits with covariates extend this simple model by allowing the reward distributions to depend upon observable side information \cite[Section 4.3]{bubeck2012regret}. For example, in sequential clinical trials the agent might have access to a patient's MRI scan or genome sequence; in web-page ad placement side-information might include a particular user's preferences and purchasing history. Owing to their widespread applicability, multi-armed bandits with covariates have been extensively studied (\cite{beygelzimer2011contextual,kakade2008efficient,langford2008epoch,perchet2013multi,qian2016kernel,rigollet2010nonparametric,seldin2011pac,slivkins2011contextual,wang2005arbitrary,wang2005bandit,yang2002randomized}). In this paper we shall consider the non-parametric setting in which the relationship between reward distribution and side-information is assumed to satisfy smoothness conditions, without specifying a particular parametric form. Yang and Zhu proved strong-consistency for an epsilon-greedy approach to this problem, using either nearest neighbour or histogram based methods to model the functional dependency of the reward distribution upon the covariate (\cite{yang2002randomized}). Rigollet and Zeevi introduced the UCBogram which partitions the covariate space into cubes and runs the UCB locally on each member of the partition (\cite{rigollet2010nonparametric}). Rigollet and Zeevi prove a regret bound with exponents depending upon distributional assumptions including a natural extension of the Tysbakov margin condition (\cite{tsybakov2004optimal}). Unfortunately, the regret bound is sub-optimal when the margin parameter is greater than one. Later Perchet and Rigollet developed the Adaptively Binned Successive Elimination algorithm (ABSE) which runs the Successive Elimination algorithm locally on increasingly refined partitions of the covariate space (\cite{perchet2013multi}). Perchet and Rigollet demonstrated that the ABSE algorithm achieves minimax optimal regret guarantees for all values of the margin parameter (\cite{perchet2013multi}). Hence, the adaptive refinement of the partition of the covariate space enables the ABSE algorithm to take advantage of low-noise conditions, expressed as a margin condition. 

Despite the strong theoretical merits of the ABSE algorithm, there are several limitations owing to its dependency upon a partition of the feature space into dyadic hyper-cubes. Firstly, there are many applications in which it is natural to construct a metric between data points which cannot be embedded in a Euclidean space without significant distortion. Examples include the Wasserstein distance between images and the edit distance on graphs, \cite{frogner2015learning,luxburg2004distance}. However, neither the UBogram nor the ABSE algorithm can be applied to non-Euclidean metric spaces. 

Secondly, the regret bounds for the ABSE algorithm require that the marginal distribution $\mu$ be Lebesgue absolutely continuous with a density bounded from below on the unit hyper-cube $[0,1]^D$. Whilst this condition is not entirely necessary for the analysis, the proof does depend crucially upon the existence of constants $C_d, d>0$ such that the following holds. For every dyadic hyper-cube $B\subset [0,1]^D$ of the form $B=2^{-q}\cdot \prod_{i=1}^D[z_i,z_i+1]$ with $z_1,\cdots,z_D,q\in \N\cup\{0\}$, we have either $\mu(B)\geq C_d\cdot \diam(B)^d$ or $\mu(B)=0$. However, this condition does not hold for many well-behaved measures on Euclidean space (see Appendix \ref{exampleDiadicSmallMeasureAppendix} for a simple example).

Thirdly, the construction of the partitions in both the UCBogram and the ABSE algorithm requires prior knowledge of the intrinsic dimensionality of the covariate space as an input parameter. In the case of the UCBogram the dimension $d$ is used to choose the optimal partition size \cite[Theorem 3.1]{rigollet2010nonparametric}. In the case of the ABSE algorithm, a partition element $B$ is refined after $l_B$ rounds, where $l_B$ is a number which depends upon the dimension $d$ \cite[Equation (5.2)]{perchet2013multi}. Moreover, if covariates are supported on a low-dimensional sub-manifold, then the intrinsic dimensionality of the sub-manifold is unlikely to be known in advance. The aim of the current paper is to address these three limitations.

The $k$-nearest neighbour method is amongst the simplest approaches to supervised learning. In addition, it has strong theoretical guarantees. Kpotufe has shown that the $k$-nearest neighbour regression algorithm attains distribution dependent minimax optimal rates, without prior knowledge of the intrinsic dimensionality of the data (\cite{kpotufe2011k}). Chaudhuri and Dasgupta have shown the $k$-nearest neighbour method attains distribution dependent minimax optimal rates in the supervised classification setting (\cite{chaudhuri2014rates}). In particular, the $k$-nearest neighbour classifier automatically takes advantage of low noise in the data, expressed as a margin condition. In light of these theoretical strengths, it is natural to apply the $k$-nearest neighbour method to problem of multi-armed bandits with covariates.

We propose the $k$-nearest neighbour UCB algorithm ($k$-NN UCB), a conceptually simple procedure for multi-armed bandits with covariates which combines the UCB algorithm with $k$-nearest neighbour regression. The algorithm does not require prior knowledge of the intrinsic dimensionality of the data. It is also naturally anytime, without resorting to the doubling trick. We prove a regret bound for the $k$-NN UCB algorithm which is minimax optimal up to logarithmic factors. In particular, the algorithm automatically takes advantage of both low intrinsic dimensionality of the marginal distribution over the covariates and low noise conditions, expressed as a margin condition. In addition, focusing on the case of bounded rewards, we give corresponding regret bounds for the $k$-nearest neighbour KL-UCB algorithm ($k$-NN KL-UCB), which is an analogue of the KL-UCB algorithm (\cite{garivier2011kl}) adapted to the setting of multi-armed bandits with covariates. Finally, we present empirical results which demonstrate the ability of both $k$-NN UCB and $k$-NN KL-UCB to take advantage of situations where the data is supported on an unknown sub-manifold of a high-dimensional feature space.

%% file: Sections/problemSetup.tex
In this section we shall introduce some notation and background.

\subsection{Notation}

We consider the problem of bandits with covariates on metric spaces. Suppose we have a metric space $\left(\X,\rho\right)$. Given $x \in \X$ and $r>0$ we let $B(x,r)$ denote the open metric ball of radius $r$, centred at $x$. Given $q \in \N$ we let $[q]=\{1,\cdots,q\}$. Given a collection of $A$ arms, we let $\Prob$ denote a distribution over random variables $(X,Y)$ with $X\in \X$ and $Y=(Y^a)_{a\in [A]} \in \R^A$, where $Y^a$ denotes the value of arm $a$. We let $\mu$ denote the marginal of $\Prob$ over $X\in \X$ and let $\supp(\mu)$ denote its support. For each $a\in [A]$ we define a function $f^a:\X\rightarrow \left[0,1\right]$ by $f^a(x)=\E\left[Y^a |X=x\right]$. 

For each $t\in [n]$ a random sample $(X_t,Y_t)$ is drawn i.i.d from $\Prob$. We are allowed to view the feature vector $X_t\sim \mu$ and we must choose an arm $a \in [A]$ and receive the stochastic reward $Y^{a}_t$. We are able to observe the value of our chosen arm, but not the value of the remaining arms. Our sequential choice of arms is given by a policy $\pi=\{\pi_t\}_{t\in [n]}$ consisting of functions $\pi_t:\X\rightarrow [A]$, where $\pi_t$ is determined purely by the known reward history $\D_{t-1}=\left\lbrace \left(X_s,\pi_s,Y^{\pi_s}_s\right)\right\rbrace_{s\in [t-1]}$. The goal is to choose $\pi_t$ so as to maximise the cumulative reward $\sum_{t\in [n]}Y_t^{\pi_t}$. In order to quantify the quality of a policy $\pi$ we compare its expected cumulative reward to the cumulative reward to that of an oracle policy $\pi^* = \left\lbrace \pi^*_t\right\rbrace_{t \in [n]}$ defined by $\pi^*_t \in \text{argmax}_{a \in [A]}\left\lbrace f^a(X_t)\right\rbrace$. We define the regret by $R_n\left(\pi\right)=\sum_{t\in [n]}\left(Y^{\pi^*_{t}}_t-Y^{\pi_t}_t\right)$.

%% file: Sections/assumptions.tex
\subsection{Assumptions}

We shall make the following assumptions:  

\begin{assumption}[Dimension assumption]\label{dimensionAssumption} There exists $C_{\Xdim},\Xdim,R_{\X}>0$ such that for all $x \in \supp(\mu)$, $r\in \left(0,R_{\mathcal{X}}\right)$ we have\\ \mbox{$ \mu\left(B(x;r)\right)\geq C_{\Xdim}\cdot r^{\Xdim}$}.
\end{assumption}

Assumption \ref{dimensionAssumption} holds for well-behaved measures $\mu$ which are absolutely continuous with respect to the Riemannian volume form $V_{\mathcal{M}}$ on a $d$-dimensional sub-manifold of Euclidean space (see Proposition \ref{manifoldImpliesDimensionAssumption}, Appendix \ref{manifoldsSec}). See Appendix \ref{exampleDiadicSmallMeasureAppendix} for an example where Assumption \ref{dimensionAssumption} whilst the measure of dyadic sub-cubes is not well behaved.

\begin{assumption}[Lipschitz assumption]\label{lipschitzAssumption} There exists a constant $\lambda>0$ such that for all $a \in [A]$, $x_0,x_1\in \X$ we have\\ $\big|f^a(x_0)-f^a(x_1)\big|\leq \lambda \cdot \rho\left(x_0,x_1\right)$.
\end{assumption}

Assumption \ref{lipschitzAssumption} quantifies the requirement that similar covariates should imply similar conditional reward expectations. Let $f^*(x)=\max_{a \in [A]}\left\lbrace f^a(x)\right\rbrace$. For each $a \in [A]$ let $\Delta^a(x)=f^*(x)-f^a(x)$, and define
\begin{align*}
\Delta(x) = \begin{cases} \min_{a \in [A]}\left\lbrace \Delta^a(x): \Delta^a(x)>0\right\rbrace \text{ if }\exists a \in [A]\hspace{2mm} \Delta^a(x)>0\\
0 \text{ otherwise. }
\end{cases}
\end{align*}

\begin{assumption}[Margin assumption]\label{marginAssumption} There exists $\delta_{\alpha},C_{\alpha},\alpha >0$ such that for all $\delta\in \left(0,\delta_{\alpha}\right)$ we have\\ $\mu\left(\left\lbrace x\in \X: 0<\Delta(x)<\delta\right\rbrace\right)\leq C_{\alpha}\cdot \delta^{\alpha}$.
\end{assumption}
 Assumption \ref{marginAssumption} quantifies the difficulty of the problem. It is a natural analogue of Tysbakov's margin condition (\cite{tsybakov2004optimal}) introduced by \cite{rigollet2010nonparametric}. Perchet and Rigollet showed that if $\X$ is a manifold and $\alpha > d$ then we must have $\eta(x)\neq \infty$ on the interior of $\supp(\mu)$ \cite[Proposition 3.1]{perchet2013multi}. All of our theoretical results require assumptions \ref{dimensionAssumption}, \ref{lipschitzAssumption} and  \ref{marginAssumption}. We shall also use one of the following two assumptions.

\begin{assumption}[Subgaussian noise assumption]\label{subgaussianNoiseAssumption} For each $t \in [n]$ and $a \in [A]$ the arms $Y^a_t$ have sub-gaussian noise ie. for all $x \in \X$ and $\theta\in \R$, 
\[\E\left[ \exp\left( \theta \cdot \left(Y_t^a-f^a(x)\right)\right)|X_t=x \right]\leq \exp\left(\theta^2/2\right).\]
\end{assumption}

\begin{assumption}[Bounded rewards assumption]\label{boundedRewardsAssumption} For all $t \in [n]$ \& $a \in [A]$, $Y_t^a \in [0,1]$.
\end{assumption}

%Suppose also that $M:=\sup\left\lbrace \Delta^a(x): x \in \X,\hspace{2mm}a \in [A]\right\rbrace<\infty$.

%% file: Sections/nearestNeighbourAlgorithms.tex
In this section we introduce a pair of nearest neighbour based UCB strategies. We begin by introducing a generalized $k$-nearest neighbours index strategy, of which the other strategies are special cases.

\subsection{The generalized k-nearest neighbours index strategy}
Suppose we are at a time step $t \in [n]$ and we have access to the reward history $\D_{t-1}$. For each $x \in \X$ we let $\left\lbrace \tau_{t,q}(x)\right\rbrace_{q\in [t-1]}$ be an enumeration of $[t-1]$ such that for each $q \leq t-2$, \[\rho\left(x,X_{\tau_{t,q}(x)}\right) \leq \rho\left(x,X_{\tau_{t,q+1}(x)}\right).\]
Given $x \in \X$ and $k \in [t-1]$ we define $\Gamma_{t,k}(x):=\left\lbrace \tau_{t,q}(x):q \in [k]\right\rbrace \subseteq [t-1]$ and let 
\begin{align*}
r_{t,k}(x)=\max\left\lbrace \rho\left(x,X_s\right):s\in \Gamma_{t,k}(x)\right\rbrace=\rho\left(x,X_{\tau_{t,k}(x)}\right).
\end{align*}
We adopt the convention that $0/0:=0$. For each $a \in [A]$ we define
\begin{align*}
N^a_{t,k}(x)&:=\sum_{s \in \Gamma_{t,k}(x)}\one\left\lbrace \pi_s=a\right\rbrace,\\
S^a_{t,k}(x)&:=\sum_{s \in \Gamma_{t,k}(x)}\one\left\lbrace \pi_s=a\right\rbrace\cdot Y^a_{s},\\
\hat{f}^a_{t,k}(x) &:= S^a_{t,k}(x)/N^a_{t,k}(x).
\end{align*}

In addition, given a constant $\theta>0$ and a non-decreasing function $\varphi:\N\rightarrow [1,\infty)$ we define a corresponding uncertainty value $U_{t,k}^{a}\left(x\right)$ by
\begin{align*}
U_{t,k}^{a}\left(x\right):=\sqrt{\left(\theta \log t\right)/N^a_{t,k}(x)}+\varphi(t) \cdot r_{t,k}(x).
\end{align*}
We shall combine $\hat{f}_{t,k}^a(x)$, $U_{t,k}^{a}\left(x\right)$, $N^a_{t,k}(x)$ and $r_{t,k}(x)$ to construct an index $\I_{t,k}^a(x)$ corresponding to an upper-confidence bound on the reward function $f^a(x)$. Our algorithm then proceeds as follows. At each time step $t$, a feature vector $X_t$ is received. For each arm $a \in [A]$, the algorithm selects a number of neighbours $k_t(a)$ by minimising the uncertainty $U_{t,k}^{a}\left(X_t\right)$. The algorithm then selects the arm which maximises the index $\I_{t,k_t(a)}^a(X_t)$. The psuedo-code for this generalised k-NN index strategy is presented in Algorithm \ref{generalisedKnnIndexAlgorithm}.
\paragraph{}

\begin{algorithm}[htbp]
\floatconts
{alg:KNNInd}
 {\caption{The k-NN index strategy \label{generalisedKnnIndexAlgorithm}}}
{ \begin{enumerate}
     \item \text{For} {$t=1,\cdots,A$,} \text{do} {$\pi_t = t$;}
 \item \text{For} {$t=A+1,\cdots,n$,}
 
 \begin{enumerate}
     \item  Observe $X_t$;
     \item \text{For }$a = 1,\cdots,A$,\\ Choose $k_t(a) \gets \argmin_{k \in [t-1]}\left\lbrace  U_{t,k}^{a}(X_t)\right\rbrace$;
     \item Choose $\pi_t \in \argmax_{a \in [A]}\left\lbrace \I_{t,k_t(a)}^a(X_t)\right\rbrace$;
     \item Receive reward $Y_t^{\pi_t}$;
 \end{enumerate}
 \end{enumerate}
 }
\end{algorithm}

\paragraph{}
By selecting $k_t(a)$ so as to minimise the $U_{t,k}^{a}\left(X_t\right)$ we avoid giving an explicit formula for $k$. This is fortuitous, since in order to obtain optimal regret bounds, any such formula would necessarily depend upon both the time horizon $n$ and the intrinsic dimensionality of the data $d$, and in general, neither $n$ nor $d$ will be known a priori by the learner. Selecting $k_t(a)$ in this way is inspired by Kpotufe's procedure for selecting $k$ in the regression setting, so as to minimise an upper bound on the squared error (\cite{kpotufe2011k}).

\subsection{k-Nearest Neighbour UCB}

The $k$-Nearest Neighbour UCB algorithm ($k$-NN UCB) is a special case of Algorithm \ref{generalisedKnnIndexAlgorithm} with the following index function,
\begin{align}\label{knnUCBIndexEquation}
\I_{t,k}^a(x)=\hat{f}^a_{t,k}(x)+U_{t,k}^a(x).
\end{align}

The $k$-NN UCB algorithm satisfies the following regret bound whenever the noise is subgaussian (Assumption \ref{subgaussianNoiseAssumption}). First we let $\varphi^{-1}(\lambda):=\inf\left\lbrace t \in \N: \varphi(t)\geq \lambda\right\rbrace$ and define $M:=\max_{a \in [A]}\left\lbrace\sup_{x \in \X}\left\lbrace\Delta^a(x)\right\rbrace\right\rbrace$. For all $n\in \N$ let $\overline{\log}(n):=\max\{1,\log(n)\}$.

\begin{restatable}{theorem}{knnUCBRegretBound}\label{knnUCBRegretBound} Suppose that Assumption \ref{dimensionAssumption} holds with constants $R_{\X},C_{\Xdim},\Xdim$, Assumption \ref{lipschitzAssumption} holds with Lipschitz constant $\lambda$, Assumption \ref{marginAssumption} holds with constants $\delta_{\alpha},C_{\alpha},\alpha >0$ and Assumption \ref{subgaussianNoiseAssumption} holds. Let $\pi$ be the $k$-NN UCB algorithm (Algorithm \ref{generalisedKnnIndexAlgorithm} with $\I_{t,k}^a$ as in equation (\ref{knnUCBIndexEquation})). Then for all $\theta>4$ there exists a constant $C$, depending solely upon $R_{\X},C_{\Xdim},\Xdim,\delta_{\alpha},C_{\alpha},\alpha$ and $\theta$ such that for all $n \in \N$ we have
\begin{align*}
\E\left[R_n(\pi)\right]&\leq M\cdot \varphi^{-1}(\lambda)+ C \cdot A \cdot \left( M \cdot \varphi(n)^d+ n \cdot \left(\frac{ \varphi(n)^d \cdot \overline{\log}(n)}{ n}\right)^{\min\left\lbrace \frac{\alpha+1}{d+2},1\right\rbrace}\right).
\end{align*}
\end{restatable}
Theorem \ref{knnUCBRegretBound} follows from the more general Theorem \ref{generalisedIndexRegretBound} in Section \ref{regretAnalysisSec}. The full proof is given in Appendix \ref{proofOfKnnUCBRegretBoundSec}. Note that by taking $\varphi(n) = O(\log n)$ we obtain a regret bound which is minimax optimal up to logarithmic factors for any smooth compact embedded sub-manifold (See Theorem \ref{lowerBound}, Appendix \ref{manifoldsSec} for details).

\subsection{k-Nearest Neighbour KL-UCB}

The $k$-Nearest Neighbour KL-UCB algorithm is another special case of Algorithm \ref{generalisedKnnIndexAlgorithm}, customized for the setting of bounded rewards. The $k$-Nearest Neighbour KL-UCB algorithm is an adaptation of the KL-UCB algorithm of \cite{garivier2011kl}, which has shown strong empirical performance combined with tight regret bounds. Given $p,q \in [0,1]$ we define the Kullback-Leibler divergence $d(p,q)$ by
\begin{align*}
d(p,q):=p \log\left(p/q\right)+\left(1-p\right)\cdot\log\left(\left(1-p\right)/\left(1-q\right)\right).
\end{align*}

\begin{align}\label{knnKlUCBIndexEquation}
\I_{t,k}^a(x)=\sup\left\lbrace \omega \in [0,1]: N_{t,k}^a(x) \cdot d\left(\hat{f}^a_{t,k}(x),\omega\right)\leq \theta \cdot \log t \right\rbrace + \varphi(t)\cdot r_{t,k}(x).
\end{align}

\begin{restatable}{theorem}{knnKlUCBRegretBound}\label{knnKlUCBRegretBound} Suppose that Assumption \ref{dimensionAssumption} holds with constants $R_{\X},C_{\Xdim},\Xdim$, Assumption \ref{lipschitzAssumption} holds with Lipschitz constant $\lambda$, Assumption \ref{marginAssumption} holds with constants $\delta_{\alpha},C_{\alpha},\alpha  >0$ and Assumption \ref{boundedRewardsAssumption} holds. Let $\pi$ be the $k$-NN KL-UCB algorithm (Algorithm \ref{generalisedKnnIndexAlgorithm} with $\I_{t,k}^a$ as in equation (\ref{knnKlUCBIndexEquation})). Then for all $\theta>2$ there exists a constant $C$, depending solely upon $R_{\X},C_{\Xdim},\Xdim,\delta_{\alpha},C_{\alpha},\alpha $ and $\theta$ such that for all $n \in \N$ we have
\begin{align*}
\E\left[R_n(\pi)\right]&\leq \varphi^{-1}(\lambda)+ C \cdot A \cdot \left(  \varphi(n)^d+ n \cdot \left(\frac{ \varphi(n)^d \cdot \overline{\log}(n)}{ n}\right)^{\min\left\lbrace \frac{\alpha+1}{d+2},1\right\rbrace}\right).
\end{align*}
\end{restatable}

Theorem \ref{knnKlUCBRegretBound} follows from the more general Theorem \ref{generalisedIndexRegretBound} in Section \ref{regretAnalysisSec}. The full proof is given in Appendix \ref{proofOfKnnKlUCBRegretBoundSec}. As with Theorem \ref{knnUCBRegretBound} we may select $\varphi(n) = O(\log n)$ to obtain a regret bound which is minimax optimal up to logarithmic factors. Experiments on synthetic data indicate that the $k$-NN KL-UCB algorithm typically outperforms the $k$-NN UCB algorithm, just as the KL-UCB (\cite{garivier2011kl}) algorithm typically outperforms the standard UCB algorithm (see Section \ref{experimentalSec}). However, the regret bounds in Theorems \ref{knnUCBRegretBound} and \ref{knnKlUCBRegretBound} are of the same order.

%% file: Sections/analysisForGeneralisedIndexAlgorithm.tex
In order to prove Theorems \ref{knnUCBRegretBound} and \ref{knnKlUCBRegretBound} we first prove the more general Theorem \ref{generalisedIndexRegretBound}. Suppose we have a k-NN index strategy (Algorithm \ref{generalisedKnnIndexAlgorithm}) with index $\I^a_{t,k}$.  We shall define for the index strategy a set of good events $\left\lbrace \mathcal{G}_t\right\rbrace_{t \in [n]}$ as follows. For each $a \in [A]$, $t \in [n]$ and $k \in [t-1]$ we define the event
\begin{align*}
\mathcal{G}_{t,k}^a:=\left\lbrace \varphi(t) \geq \lambda \right\rbrace \cap \left\lbrace \I^a_{t,k}(X_t) - 2\cdot U_{t,k}^a(X_t) \leq f^a(X_t) \leq \I^a_{t,k}(X_t) \right\rbrace.
\end{align*}
Let $\mathcal{G}_t:= \bigcap_{a \in [A]}\bigcap_{k \in [t-1]}\mathcal{G}_{t,k}^a$.

\begin{restatable}{theorem}{generalisedIndexRegretBound}\label{generalisedIndexRegretBound} Suppose that Assumption \ref{dimensionAssumption} holds with constants $R_{\X},C_{\Xdim},\Xdim$, Assumption \ref{lipschitzAssumption} holds with Lipschitz constant $\lambda$ and Assumption \ref{marginAssumption} holds with constants $\delta_{\alpha},C_{\alpha},\alpha >0$. Suppose $\pi$ is a $k$-NN index strategy (Algorithm \ref{generalisedKnnIndexAlgorithm}) with index $\I_{t,k}^a$. Then there exists a constant $C$, depending solely upon $R_{\X},C_{\Xdim},\Xdim,\delta_{\alpha},C_{\alpha},\alpha $ such that for all $n \in \N$ we have
\begin{align*}
\E\left[R_n(\pi)\right]&\leq C \cdot A \cdot \left( M\cdot \varphi(n)^d+n \cdot \left(\frac{\theta \cdot \varphi(n)^d \cdot \overline{\log}(n)}{ n}\right)^{\min\left\lbrace \frac{\alpha+1}{d+2},1\right\rbrace}\right)\\&\hspace{1cm}+M\cdot\sum_{t\in [n]}\left(1-\Prob\left[\mathcal{G}_{t}\right]\right).
\end{align*}
\end{restatable}

Theorems \ref{knnUCBRegretBound} and \ref{knnKlUCBRegretBound} are deduced from Theorem \ref{generalisedIndexRegretBound} in Appendices \ref{proofOfKnnUCBRegretBoundSec} and \ref{proofOfKnnKlUCBRegretBoundSec}, respectively. In both cases, the deduction amounts to using concentration inequalities to show that the good events $\mathcal{G}_t$ hold with high probability. The proof of Theorem \ref{generalisedIndexRegretBound} consists of two primary components. Firstly, we prove an upper bound on the number of times an arm is pulled with covariates in a given region of the metric space with a sufficiently high local margin (see Lemma \ref{numberOfBadPullsBound}). A key difference with the regret bounds of (\cite{rigollet2010nonparametric}, \cite{perchet2013multi}) is that these local bounds hold for arbitrary subsets, rather than just the members of the partition constructed by the algorithm. Secondly, we construct a partition of the covariate space based on local values of the margin, with regions of low margin partitioned into smaller pieces (see the proof of Proposition \ref{regretOneArmLemma}). The local upper bound is then applied to members of the partition to derive the regret bound.

Given a subset $B\subseteq \X$ and $a \in [A]$ we define $\Delta^a(B):= \sup_{x \in B}\left\lbrace \Delta^a(x)\right\rbrace$ and let
\begin{align*}
T_n^a\left(\pi,B\right) & := \sum_{t\in [n]} \one\left\lbrace \mathcal{G}_t\right\rbrace \cdot \one\left\lbrace X_t \in B \right\rbrace \cdot \one\left\lbrace \pi_t=a\right\rbrace\\
\tilde{R}_n^a\left(\pi,B\right) & := \sum_{t\in [n]}\one\left\lbrace \mathcal{G}_t\right\rbrace \cdot \one\left\lbrace X_t \in B \right\rbrace \cdot \one\left\lbrace \pi_t=a\right\rbrace \cdot \left(Y^{\pi^*_{t}}_t-Y^{\pi_t}_t\right).
\end{align*}

\begin{restatable}{myLemma}{breakRegretIntoSeparateArmsLemma}\label{breakRegretIntoSeparateArmsLemma} $\E\left[R_n(\pi)\right] \leq \sum_{a \in [A]}\E\left[\tilde{R}_n^a(\pi,\X)\right]+M\cdot\sum_{t\in [n]}\left(1-\Prob\left[\mathcal{G}_{t}\right]\right)$.
\end{restatable}
\begin{proof}
See Appendix \ref{localRegretSec}.
\end{proof}

In light of Lemma \ref{breakRegretIntoSeparateArmsLemma}, in order to prove Theorem \ref{generalisedIndexRegretBound} it suffices to prove the following proposition (Proposition \ref{regretOneArmLemma}).

\begin{restatable}{prop}{regretOneArmLemma}\label{regretOneArmLemma} There exists a constant $C$, depending solely upon $R_{\X},C_{\Xdim},\Xdim,\delta_{\alpha},C_{\alpha}$, \mbox{$\alpha>0$} such that for all $n \in \N$ we have
\begin{align*}
\E\left[\tilde{R}_n^a(\pi,\X)\right] \leq C \cdot  \left( M \cdot \varphi(n)^d+ n \cdot \left(\frac{\theta \cdot \varphi(n)^d \cdot \overline{\log}(n)}{ n}\right)^{\min\left\lbrace \frac{\alpha+1}{d+2},1\right\rbrace}\right).
\end{align*}
\end{restatable}

Before proving Proposition \ref{breakRegretIntoSeparateArmsLemma} we require three lemmas (\ref{regretAndCounts}, \ref{numberOfBadPullsBound} and \ref{nestedMetricCubesProp} below).

\begin{restatable}{myLemma}{regretAndCounts}\label{regretAndCounts} For any subset $B \subseteq \X$ and any $a \in [A]$ we have $\E\left[\tilde{R}_n^a(\pi,B)\right]\leq \Delta^a(B) \cdot \E\left[T^a_n(\pi,B)\right]$.
\end{restatable}
\begin{proof}
See Appendix \ref{localRegretSec}.
\end{proof}

The following key lemma bounds the number of times an arm is pulled in a given region of the covariate space.

\begin{myLemma}\label{numberOfBadPullsBound} Given a subset $B \subseteq \X$ and an arm $a \in [A]$ with \mbox{$4 \cdot \varphi(n)\cdot \diam(B)<\Delta^a(B)$}, the following holds almost surely
\begin{align*}
T_n^a(\pi,B)\leq \frac{4\theta \cdot \overline{\log n}}{\left(\Delta^a(B)-4 \cdot \varphi(n)\cdot \diam(B)\right)^2}+1.
\end{align*}
\end{myLemma}
\begin{proof}
Clearly we can assume that $T_n^a(\pi,B)>1$. We define 
\begin{align*}
t&:=\max\left\lbrace s \in [n]: X_s \in B,\hspace{1mm} \pi_s=a,\hspace{1mm} \mathcal{G}_s\text{ holds}\right\rbrace\\
k(B)&:=\max\left\lbrace q \in [t-1]:X_{\tau_{t,q}\left(X_{t}\right)} \in B\right\rbrace.
\end{align*}
Note that as $\mathcal{G}_t$ holds we must have $\lambda \leq \varphi(t) \leq \varphi(n)$. Since $X_{t} \in B$ and $X_{\tau_{t,k(B)}\left(X_{t}\right)} \in B$ we must have $r_{t,k(B)}(X_t)\leq \diam(B)$. Moreover, given any $s \in [t-1]$ with $X_s \in B$ we must have $\tau_{t,q}\left(X_{t}\right)=s$ for some $q \leq k(B)$. Thus, $T_n^a(\pi,B)\leq N_{t,k(B)}^a(X_t)+1$.

Note that $T_n^a(\pi,B)>1$ implies $t>A$. Choose $z_* \in [A]$ so that $f^{z_*}(X_t)=f^*(X_t)$. Since $\pi_t = a$ and $t>A$ we have $\mathcal{I}^{z_*}_{t,k_t({z_*})}(X_t)\leq \mathcal{I}^{a}_{t,k_t({a})}(X_t)$. On the other hand, since $\mathcal{G}_t$ holds we have,
$f^{z_*}(X_t) \leq \mathcal{I}^{z_*}_{t,k_t({z_*})}(X_t)$ and 
\begin{align*}
f^{a}(X_t) \geq \mathcal{I}^{a}_{t,k_t({a})}(X_t)-2\cdot U_{t,k_t({a})}^{a}(X_t) 
\end{align*}
Thus, given above and the definitions of $k_t(a)$ and $U^a_{t,k}$ we have
\begin{align*}
\left(f^{z_*}(X_t)-f^a(X_t)\right)/2 &\leq   U^a_{t,k_t(a)}(X_t)\leq  U^a_{t,k(B)}(X_t)\\
& =\sqrt{{(\theta \log t)}/N_{t,k(B)}^a(X_t)}+ \varphi(t) \cdot r_{t,k(B)}(X_t)\\
& \leq \sqrt{{(\theta \log t)}/{(T_n^a(\pi,B)-1)}}+ \varphi(n) \cdot \diam(B).
\end{align*}
By the Lipschitz assumption (Assumption \ref{lipschitzAssumption}) together with the fact that $X_t \in B$ we must have \[f^{z_*}(X_t)-f^a(X_t) \geq \Delta^a(B)-2 \lambda \cdot \diam(B)\geq  \Delta^a(B)-2 \varphi(n) \cdot \diam(B).\] Combining with the above proves the lemma.
\end{proof}

Lemma \ref{nestedMetricCubesProp} applies Assumption \ref{dimensionAssumption} to obtain an analogue of nested hyper-cubes within $[0,1]^d$. The proof adapts ideas from geometric measure theory (\cite{kaenmaki2012existence}).
\begin{restatable}{myLemma}{nestedMetricCubesProp}\label{nestedMetricCubesProp} Suppose that Assumption \ref{dimensionAssumption} holds. Given $q \in \mathbb{N}\backslash\{0\}$, $\delta \in \left(0,R_{\X}\right]$ and $r\in \left(0,1/3\right)$ there exists a finite  collection of subsets $\left\lbrace Z_{l,i}:l \in [q],i \in [m_l]\right\rbrace$ which satisfies:
\begin{enumerate}
\item For each $l \in [q]$, $\left\lbrace Z_{l,i}\right\rbrace_{i \in [m_l]}$ is a partition of $\X$.
\item Given $l_1,l_2 \in [q]$ with $l_1 \leq l_2$, $i_1 \in [m_{l_1}]$ and $i_2 \in [m_{l_2}]$, either \mbox{$Z_{l_1,i_1}\cap Z_{i_2,l_2}= \emptyset$} or $Z_{l_2,i_2}\subseteq Z_{l_1,i_1}$.
\item For all $l \in [q]$, $i \in [m_l]$ we have $\diam(Z_{l,i})\leq \delta\cdot r^l$ and 
\begin{align*}
\mu\left(Z_{l,i}\right) \geq C_{\Xdim}\cdot \left( (\delta/4)\cdot (1-3r)\cdot r^l\right)^d.
\end{align*}
\end{enumerate}
\end{restatable}
\begin{proof}
See Appendix \ref{nestedPartitionSec}.
\end{proof}
We are now ready to complete the proof of Proposition \ref{regretOneArmLemma}, which entails Theorem \ref{generalisedIndexRegretBound}.
\begin{proof}[Proof of Proposition \ref{regretOneArmLemma}]
Throughout the proof $c_1,\cdots, c_7$ will denote constants depending solely upon $R_{\X},C_{\Xdim},\Xdim,\delta_{\alpha},C_{\alpha},\alpha$. We shall apply Lemma \ref{nestedMetricCubesProp} to construct a cover of $\X$ based upon the local value of $\Delta^a$. First let $\delta(n):=\min\left\lbrace R_{\X},\delta_{\alpha}/(10 \cdot \varphi(n))\right\rbrace$. Take some $q \in \N$ (to be specified later), let $\delta=\delta(n)$ and $r=1/4$ and let $\left\lbrace Z_{l,i}: l \in [q], i \in [m_l]\right\rbrace$ be a collection of subsets satisfying properties (1),(2),(3) from Lemma \ref{nestedMetricCubesProp}. In particular, for all $l \in [q]$ and $i \in [m_l]$ we have $\diam(Z_{l,i}) \leq \delta(n) \cdot 4^{-l}$ and $\mu(Z_{l,i})\geq C_d \cdot (\delta(n)/16)^d \cdot 4^{-ld}$. First let 
\begin{align*}
\mathcal{Z}^a_{\text{big}}:=\left\lbrace Z_{1,i}:i\in [m_1],\hspace{2mm} \Delta^a(Z_{1,i})\geq 5\cdot \varphi(n) \cdot \delta(n)\right\rbrace.
\end{align*}
For each $l \in [q]$ we define 
\begin{align*}
\mathcal{Z}^a_l:=\left\lbrace Z_{l,i}:i\in [m_l],\hspace{2mm} 5\cdot \varphi(n) \cdot \delta(n)\cdot 4^{-l} \leq \Delta^a(Z_{l,i}) < 5\cdot \varphi(n) \cdot \delta(n)\cdot 4^{-l+1}\right\rbrace.
\end{align*}
Finally, define 
\begin{align*}
Z^a_{\text{small}}&:= \left\lbrace x \in \X:0<\Delta^a(x)<5\cdot \varphi(n) \cdot \delta(n)\cdot 4^{-q}\right\rbrace\\
Z^a_0&:=\left\lbrace x\in \X:\Delta^a(x)=0\right\rbrace.
\end{align*}
We claim that for all $r\in [q]$ we have
\begin{align*}
\mathcal{X}\subseteq \bigcup \left(\mathcal{Z}^a_{\text{big}} \cup \left(\bigcup_{l\in [r]}\mathcal{Z}^a_l\right)\cup\left\lbrace Z_{r,i}:i \in [m_{r}],\hspace{2mm}\Delta^a(Z_{r,i})<5\cdot \varphi(n) \cdot \delta(n)\cdot 4^{-r}\right\rbrace\right).
\end{align*}
For $r=1$ the claim follows straightforwardly from the fact that $\left\lbrace Z_{1,i}\right\rbrace_{i \in [m_1]}$ is a partition of $\X$. Now suppose the claim holds for some $r\in[q-1]$. By properties (1) and (2) in Lemma \ref{nestedMetricCubesProp} for any $i \in [m_r]$, 
\begin{align*}
Z_{r,i}=\bigcup \left\lbrace Z_{r+1,j}: j \in [m_{r+1}], \hspace{1mm} Z_{r+1,j} \subseteq Z_{r,i}\right\rbrace.
\end{align*}
Moreover, if $Z_{r+1,j}\subseteq Z_{r,i}$ then $\Delta^a\left(Z_{r+1,j}\right)\leq \Delta^a\left(Z_{r,i}\right)$. Thus, we have
\begin{align*}
\bigcup & \left\lbrace Z_{r,i}:i \in [m_{r}],\hspace{2mm}\Delta^a(Z_{r,i})<5\cdot \varphi(n) \cdot \delta(n)\cdot 4^{-r}\right\rbrace\\
&\subseteq \bigcup  \left\lbrace Z_{r+1,i}:i \in [m_{r+1}],\hspace{2mm}\Delta^a(Z_{r+1,i})<5\cdot \varphi(n) \cdot \delta(n)\cdot 4^{-r}\right\rbrace\\
&=\bigcup \left(\mathcal{Z}_{r+1}^a\cup \left\lbrace Z_{r+1,i}:i \in [m_{r+1}],\hspace{2mm}\Delta^a(Z_{r+1,i})<5\cdot \varphi(n) \cdot \delta(n)\cdot 4^{-r-1}\right\rbrace\right).
\end{align*}
Hence, given that the claim holds for $r$ it must also hold for $r+1$. 
From the special case where $r=q$ we deduce that,
\begin{align*}
\mathcal{X}\subseteq \bigcup \left(\mathcal{Z}^a_{\text{big}} \cup \left(\bigcup_{l\in [q]}\mathcal{Z}^a_l\right)\cup\left\lbrace Z_{\text{small}}^a,Z_0^a\right\rbrace\right).
\end{align*}
Thus, given that $\E\left[\tilde{R}_n^a\left(\pi,Z_{0}^a\right)\right]=0$ we have
\begin{align*}
\E\left[\tilde{R}_n^a\left(\pi,\X\right)\right]&\leq \sum_{Z \in \mathcal{Z}^a_{\text{big}}} \E\left[\tilde{R}_n^a\left(\pi,Z\right)\right]+\sum_{l=1}^q\sum_{Z \in \mathcal{Z}^a_{l}} \E\left[\tilde{R}_n^a\left(\pi,Z\right)\right]+\E\left[\tilde{R}_n^a\left(\pi,Z_{\text{small}}^a\right)\right].
\end{align*}
We begin by considering $\sum_{Z \in \mathcal{Z}^a_{\text{big}}}\E\left[ \tilde{R}_n^a\left(\pi,Z\right)\right]$. Given $Z \in \mathcal{Z}^a_{\text{big}}$ we have
\mbox{$\diam(Z) \leq \delta(n)/4$}, $5\cdot \varphi(n) \cdot \delta(n) \leq \Delta^a(Z) \leq M$ and $\mu(Z)\geq C_d \cdot (\delta(n)/64)^d$. By Lemmas \ref{regretAndCounts} and \ref{numberOfBadPullsBound} we have
\begin{align*}
\E\left[\tilde{R}_n^a\left(\pi,Z\right)\right] &\leq \Delta^a(Z) \cdot \left( \frac{4\theta \cdot \overline{\log}(n)}{\left(\Delta^a(Z)-4\cdot \varphi(n) \cdot \diam(Z)\right)^2} +1\right)\\
&\leq \frac{5\theta \cdot \overline{\log}(n)}{4 \cdot \varphi(n) \cdot \delta(n)}+M. 
\end{align*}

Moreover, since $\mu(Z)\geq C_d \cdot (\delta(n)/64)^d$ for  $Z \in \mathcal{Z}^a_{\text{big}}$, we have $\#\mathcal{Z}^a_{\text{big}}\leq C_d^{-1}\cdot (\delta(n)/64)^{-d}$. Hence,
\begin{align}\label{bigMarginRegret}
\sum_{Z \in \mathcal{Z}^a_{\text{big}}} \E\left[\tilde{R}_n^a\left(\pi,Z\right)\right] \leq c_1 \cdot \varphi(n)^d \cdot \left(\theta \cdot \overline{\log}(n)+M\right).
\end{align}
Now take $l \in [q]$ and consider $Z \in \mathcal{Z}_l^a$. We have $\diam(Z) \leq \delta(n) \cdot 4^{-l}$,
\[5 \cdot  \varphi(n) \cdot \delta(n) \cdot 4^{-l} \leq \Delta^a(Z) < 5\cdot  \varphi(n) \cdot \delta(n) \cdot 4^{-l+1}\] 
and $\mu(Z)\geq C_d \cdot (\delta(n)/16)^d \cdot 4^{-ld}$. Hence, by Lemma \ref{numberOfBadPullsBound} we have 
\begin{align*}
T_n^a\left(\pi,Z\right)\leq \frac{\theta \cdot \overline{\log}(n)}{\left(\varphi(n)\cdot \delta(n)\right)^2}\cdot 4^{2l+1}+1.
\end{align*}
Combining with Lemma \ref{regretAndCounts} and $\Delta^a(Z) < 5\cdot  \varphi(n) \cdot \delta(n) \cdot 4^{-l+1}$ we have
\begin{align*}
\E\left[\tilde{R}_n^a\left(\pi,Z\right)\right] \leq c_2 \cdot \theta \cdot \overline{\log}(n) \cdot 4^l.
\end{align*}
Moreover, it follows from the definition of $\delta(n)$ that for all  $Z \in \mathcal{Z}_l^a$ we have $\Delta^a(Z)< \delta_{\alpha}$. Hence, by Assumption \ref{marginAssumption} we have
\begin{align*}
\#\mathcal{Z}^a_l \cdot C_d \cdot (\delta(n)/16)^d \cdot 4^{-ld} \leq \sum_{Z \in \mathcal{Z}^a_l}\mu\left(Z\right) \leq C_{\alpha}\cdot \left(5\cdot  \varphi(n) \cdot \delta(n) \cdot 4^{-l+1}\right)^{\alpha}.
\end{align*}
Thus, we have
\begin{align}\label{mediumMarginRegret}
\sum_{Z \in \mathcal{Z}^a_{l}} \E\left[\tilde{R}_n^a\left(\pi,Z\right)\right] \leq c_3 \cdot  \varphi(n)^d \cdot \theta \cdot \overline{\log}(n) \cdot 4^{l(d+1-\alpha)}.
\end{align}
Finally, $\Delta(Z^a_{\text{small}})\leq 5\cdot \varphi(n) \cdot \delta(n)\cdot 4^{-q}$. Hence, by Assumption \ref{marginAssumption} we have $\mu\left(Z^a_{\text{small}}\right)\leq C_{\alpha}\cdot (5\cdot \varphi(n) \cdot \delta(n))^{\alpha}\cdot 4^{-q\alpha}$. Hence, by Lemma \ref{regretAndCounts} we have 
\begin{align}\label{smallMarginRegret}
\E\left[\tilde{R}_n^a\left(\pi,Z^a_{\text{small}}\right)\right]&\leq \left(5\cdot \varphi(n) \cdot \delta(n)\cdot 4^{-q}\right) \cdot \E\left[T_n^a\left(\pi,Z^a_{\text{small}}\right)\right]\nonumber \\
&\leq \left(5\cdot \varphi(n) \cdot \delta(n)\cdot 4^{-q}\right) \cdot n \cdot \mu\left(Z^a_{\text{small}}\right) \leq c_4 \cdot n \cdot 4^{-q(\alpha+1)}.
\end{align}

Combining equations (\ref{bigMarginRegret}), (\ref{mediumMarginRegret}) and (\ref{smallMarginRegret}) we have
\begin{align*}
\E\left[\tilde{R}_n^a\left(\pi,\X\right)\right]&\leq c_5 \cdot \left(\varphi(n)^d\left(M+\theta \cdot \overline{\log}(n)\cdot \sum_{l=0}^q4^{l(d+1-\alpha)}\right)+n\cdot 4^{-q(\alpha+1)}\right)\\
& \leq c_6 \cdot  \left(\varphi(n)^d\left(M+\theta \cdot \overline{\log}(n)\cdot (1+4^{q(d+1-\alpha)})\right)+n\cdot 4^{-q(\alpha+1)}\right).
\end{align*}
Thus, if we take $q = \lceil \log\left(n/\left(\theta \cdot \varphi(n)^d \cdot \overline{\log}(n)\right)\right)/\left((d+2)\log 4\right)\rceil$ we have 
\begin{align*}
\E\left[\tilde{R}_n^a\left(\pi,\X\right)\right]&\leq c_7 \cdot \left( (M+\theta\cdot \overline{\log}(n)) \cdot \varphi(n)^d+ n \cdot \left(\frac{\theta \cdot \varphi(n)^d \cdot \overline{\log}(n)}{ n}\right)^{\frac{\alpha+1}{d+2}}\right)\\
&\leq c_8 \cdot \left( M \cdot \varphi(n)^d+ n \cdot \left(\frac{\theta \cdot \varphi(n)^d \cdot \overline{\log}(n)}{ n}\right)^{\min\left\lbrace \frac{\alpha+1}{d+2},1\right\rbrace}\right)\\
\end{align*}

\end{proof}

%% file: Sections/experimentalResults.tex
In this section we present an empirical illustration of the ability of both the $K$-NN UCB algorithm and the $K$-NN KL-UCB to adapt to the intrinsic dimensionality of the data. We consider four bandit scenarios. In each bandit scenario the marginal distribution over the covariates is supported on a $d=2$ dimensional affine sub-manifold within $\R^D$. We vary the dimension of the ambient feature space $D$ over the four scenarios so $D \in \{2,5,10,15\}$. We compare four algorithms: The UCBogram (\cite{rigollet2010nonparametric}), the ABSE algorithm (\cite{perchet2013multi}), the K-NN UCB algorithm and the K-NN KL-UCB algorithm. For further details on experimental procedure and the generation of the synthetic data we refer to Appendix \ref{experimentalDetailsSec}.

\begin{figure}
\caption{A comparison of four algorithms for multi-armed bandits with covariates: the UCBogram, the ABSE algorithm, the KNN UCB algorithm and the KNN KL-UCB algorithm. In each experiment the covariates are supported on a $d=2$-dimensional sub-manifold $\mathcal{M}\subset \R^D$ with the dimension of the ambient space varied $D\in \{2,5,10,15\}$. For each algorithm in each scenario we plot the mean and standard deviation over fifty runs. For further discussion see Section \ref{experimentalSec} \label{experimentalResultsFigure}.}
\centering
\hspace{5mm}
\includegraphics[width=1.0\textwidth]{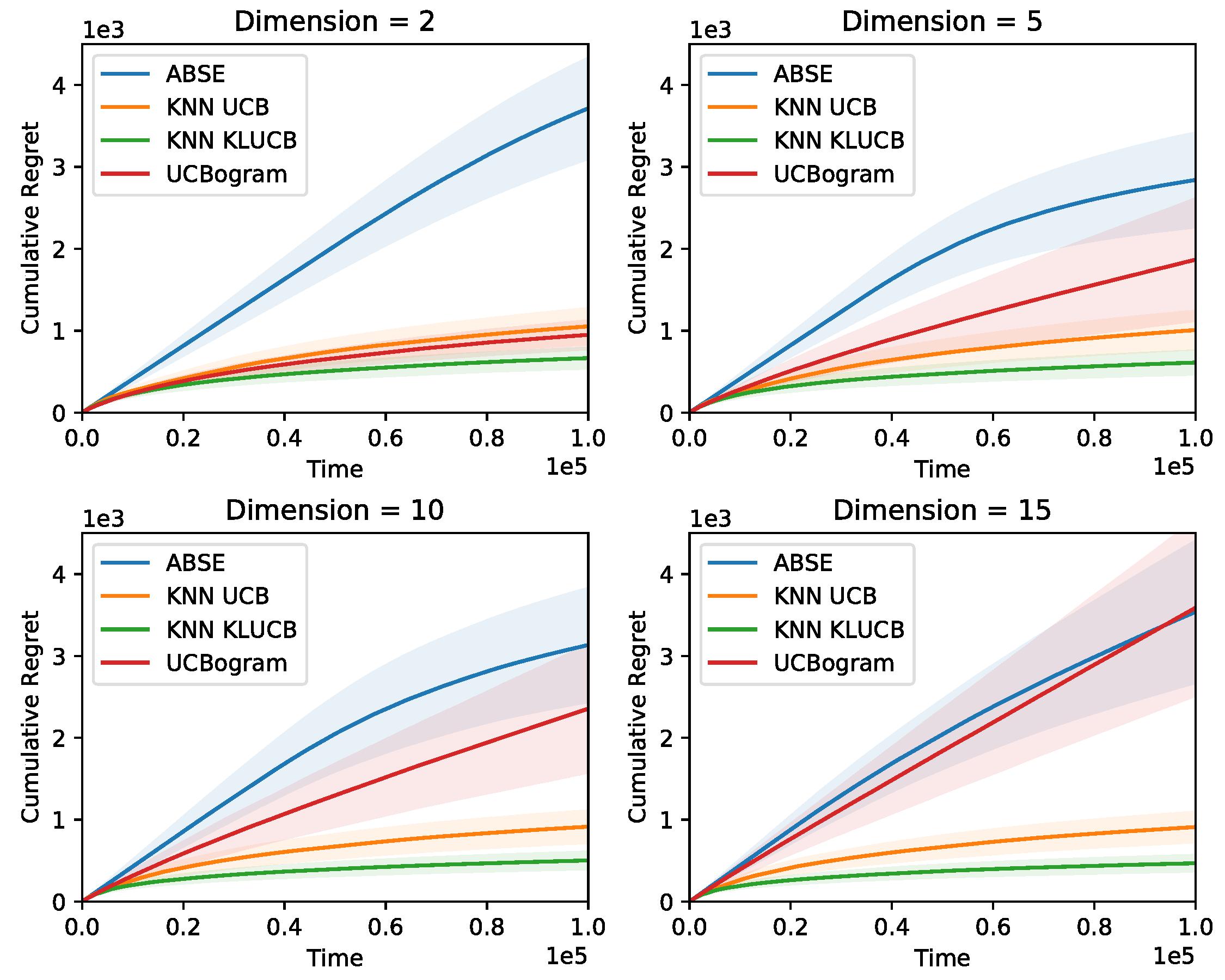}
\end{figure}

The results are displayed in Figure \ref{experimentalResultsFigure}. When $d=D=2$ the UCBogram performs comparably with the KNN based algorithms. However, the performance of the UCBogram deteriorates as we increase the dimension of the ambient feature space $D$. However, both the KNN UCB and the KNN KL-UCB algorithm are robust to increases in the ambient dimension $D$, significantly outperforming both the ABSE algorithm and the UCBogram when $D=15$. This gives an empirical illustration of the fact that bounds in Theorem \ref{knnUCBRegretBound} and \ref{knnKlUCBRegretBound} do not depend upon the dimensionality of ambient feature space.

%% file: Sections/discussion.tex
We have presented the $k$-NN UCB algorithm for multi-armed bandits with covariates. The method is conceptually simple, and unlike previous methods, such as the UCBogram and Adaptively Binned Successive Elimination, the $k$-NN UCB algorithm does not require prior knowledge of either the time horizon or the intrinsic dimension of the marginal distribution over the covariates. We have proven two regret bounds. Theorem \ref{knnUCBRegretBound} demonstrates that the $k$-NN UCB algorithm is minimax optimal, up to logarithmic factors whenever the noise is subgaussian. Theorem \ref{knnKlUCBRegretBound} demonstrates that the $k$-NN KL-UCB algorithm is minimax optimal, up to logarithmic factors, in the bounded rewards setting. Overall, we see that both $k$-NN based algorithms automatically take advantage of both low intrinsic dimensionality of the marginal distribution over the covariates and low noise in the data, expressed as a margin condition. In addition we have illustrated the robustness of the $k$-NN based algorithms to the dimension of the ambient feature space with experimental results on synthetic data.

A challenging open question remains. Is it possible to obtain algorithms which are adaptive to an unknown H\"{o}lder exponent? Both the $k$-NN UCB and the $k$-NN KL-UCB algorithm may be straightforwardly adapted to fixed H\"{o}lder exponents $\beta<1$ (in place of the Lipschitz assumption), provided the exponent is known a priori. However, it remains to be seen whether or not it is possible to construct an algorithm which automatically adapts to an unknown H\"{o}lder exponent.

\begin{comment}

 Firstly, it would be interesting to see if our regret analysis applies to Contextual-Zooming algorithm of \cite{slivkins2014contextual} in the finite arm setting. \cite[Theorem 6]{slivkins2014contextual} implies a regret bound of order $n^{1-\frac{1}{d+2}}$ in the absence of the margin condition. It would be interesting to see if a variant of contextual zooming 

\end{comment}

%% file: Sections/acknowledgements.tex
The authors gratefully acknowledge the support of the EPSRC for the LAMBDA project (EP/N035127/1) and the Manchester Centre for Doctoral Training in Computer Science (EP/1038099/1). We would also like to thank Ata Kab\'{a}n, Peter Tino, Jeremy Wyatt, Konstantinos Sechidis, Nikos Nikolaou and Sarah Nogueira for useful discussions. We would also like to thank the anonymous reviewers for their careful feedback.

%% file: Sections/knnUCBSubgaussianRegretBound.tex
 In this section we will deduce Theorem \ref{knnUCBRegretBound} from Theorem \ref{generalisedIndexRegretBound}.

 \knnUCBRegretBound*
 \begin{myLemma}\label{applyConcentrationToKnnUCBLemma}
 Suppose that Assumption \ref{lipschitzAssumption} holds with Lipschitz constant $\lambda$ and Assumption \ref{subgaussianNoiseAssumption} holds. Let $\pi$ be the $k$-NN UCB algorithm (Algorithm \ref{generalisedKnnIndexAlgorithm} with $\I_{t,k}^a$ as in equation (\ref{knnUCBIndexEquation})). Then for all $\theta>0$, all $a \in [A]$, $t \in \{\varphi^{-1}(\lambda),\cdots,n\}$ and $k \in [t-1]$ we have
 \begin{align*}
 1-\Prob\left[\mathcal{G}_{t,k}^a\right]\leq 2\cdot e \cdot \lceil (\theta/2) \cdot \left(\log t\right)^2 \rceil \cdot t^{-\theta/2}.
 \end{align*}
\end{myLemma}
\begin{proof} Recall that 
\begin{align*}
\mathcal{G}_{t,k}^a:=\left\lbrace \varphi(t) \geq \lambda \right\rbrace \cap \left\lbrace \I^a_{t,k}(X_t) - 2\cdot U_{t,k}^a(X_t) \leq f^a(X_t) \leq \I^a_{t,k}(X_t) \right\rbrace.
\end{align*}
Hence, for $t \geq \varphi^{-1}(\lambda)$, if $\mathcal{G}_{t,k}^a$ does not hold then
\begin{align*}
\left| \hat{f}^a_{t,k}(X_t)-f^a(X_t)\right|>U^a_{t,k}(X_t) &= \sqrt{\left(\theta \log t\right)/N^a_{t,k}(X_t)}+\varphi(t) \cdot r_{t,k}(X_t)\\
&\geq \sqrt{\left(\theta \log t\right)/N^a_{t,k}(X_t)}+\lambda \cdot r_{t,k}(X_t).
\end{align*}
For $s \in [t-1]$ we define 
\begin{align*}
\epsilon_s &= \one\left\lbrace s\in \Gamma_{t,k}(X_t)\right\rbrace \cdot \one\left\lbrace \pi_s=a\right\rbrace\\
Z_s & = \one\left\lbrace s\in \Gamma_{t,k}(X_t)\right\rbrace \cdot \left(Y^a_s-f^a(X_s)\right).
\end{align*}
Hence, $N^a_{t,k}(X_t)=\sum_{s \in [t-1]}\epsilon_s$. By the Lipschitz property (Assumption \ref{lipschitzAssumption}) for all $s \in \Gamma_{t,k}(X_t)$ we have 
\begin{align*}
\left|f^a(X_s)-f^a(X_t)\right|\leq \lambda \cdot \rho\left(X_s,X_t\right) \leq \lambda \cdot r_{t,k}(X_t).
\end{align*}
Thus, if $\mathcal{G}_{t,k}^a$ does not hold then
\begin{align*}
\left|\sum_{s \in [t-1]}\epsilon_s \cdot Z_s\right|>\sqrt{\left(\theta \log t\right)\cdot N^a_{t,k}(X_t)}.
\end{align*}
By Corollary \ref{subGaussianConcentrationInequality}, Appendix \ref{concentrationInequalitySec}, for any given $\left\lbrace X_s\right\rbrace_{s \in [n]}$ we have
\begin{align*}
\Prob\left[\sum_{s \in [t-1]}\epsilon_s \cdot Z_s>\sqrt{\left(\theta \log t\right)\cdot N^a_{t,k}(X_t)} \hspace{2mm}\big| \left\lbrace X_s\right\rbrace_{s \in [n]}\right]\leq e \cdot \lceil (\theta/2) \cdot \left(\log t\right)^2 \rceil \cdot t^{-\theta/2}.
\end{align*}
By the law of total expectation this implies
\begin{align*}
\Prob\left[\sum_{s \in [t-1]}\epsilon_s \cdot Z_s>\sqrt{\left(\theta \log t\right)\cdot N^a_{t,k}(X_t)} \hspace{2mm}\right]\leq e \cdot \lceil (\theta/2) \cdot \left(\log t\right)^2 \rceil \cdot t^{-\theta/2}.
\end{align*}
By applying Corollary \ref{subGaussianConcentrationInequality} to $\left\lbrace -Z_s\right\rbrace_{s \in [t-1]}$ we also have the lower tail inequality. Hence the lemma holds.
\end{proof}

\begin{proof}[Proof of  Theorem \ref{knnUCBRegretBound}] By Lemma \ref{applyConcentrationToKnnUCBLemma} we have
\begin{align*}
\sum_{t \in [n]}\left(1-\Prob\left[\mathcal{G}_t\right]\right)&\leq \varphi^{-1}(\lambda)+\sum_{a \in [A]}\sum_{t=\varphi^{-1}(\lambda)}^n\sum_{k \in [t-1]}\left(1-\Prob\left[\mathcal{G}_{t,k}^a\right]\right)\\
&\leq \varphi^{-1}(\lambda)+2e \cdot \sum_{a \in [A]}\sum_{t=\varphi^{-1}(\lambda)}^n\sum_{k \in [t-1]} \lceil (\theta/2) \cdot \left(\log t\right)^2 \rceil \cdot t^{-\theta/2}\\
&\leq \varphi^{-1}(\lambda)+2e \cdot A \cdot \sum_{t=1}^{\infty}\lceil (\theta/2) \cdot \left(\log t\right)^2 \rceil \cdot t^{1-\theta/2}.
\end{align*}
Given $\theta>4$ we have $\sum_{t=1}^{\infty}\lceil (\theta/2) \cdot \left(\log t\right)^2 \rceil \cdot t^{1-\theta/2}<\infty$. Hence, by applying Theorem \ref{generalisedIndexRegretBound}, the regret bound in Theorem \ref{knnUCBRegretBound} holds.
\end{proof}

%% file: Sections/knnKlUCBRegretBound.tex
 In this section we will deduce Theorem \ref{knnKlUCBRegretBound} from Theorem \ref{generalisedIndexRegretBound}.

 \knnKlUCBRegretBound*
 \begin{myLemma}\label{applyConcentrationToKnnKlUCBLemma}
 Suppose that Assumption \ref{lipschitzAssumption} holds with Lipschitz constant $\lambda$ and Assumption \ref{boundedRewardsAssumption} holds. Let $\pi$ be the $k$-NN KL-UCB algorithm (Algorithm \ref{generalisedKnnIndexAlgorithm} with $\I_{t,k}^a$ as in equation (\ref{knnKlUCBIndexEquation})). Then for all $\theta>0$, all $a \in [A]$, $t \in \{\varphi^{-1}(\lambda),\cdots,n\}$ and $k \in [t-1]$ we have
 \begin{align*}
 1-\Prob\left[\mathcal{G}_{t,k}^a\right]<2\cdot e \cdot \lceil \theta \cdot \left(\log t\right)^2 \rceil \cdot t^{-\theta}.
 \end{align*}
\end{myLemma}
\begin{proof}
Recall that in the $k$-NN KL-UCB algorithm we have
\begin{align*}
\I_{t,k}^a(x)=\sup\left\lbrace \omega \in [0,1]: N_{t,k}^a(x) \cdot d\left(\hat{f}^a_{t,k}(x),\omega\right)\leq \theta \cdot \log t \right\rbrace + \varphi(t)\cdot r_{t,k}(x).
\end{align*}
For the purposes of the proof we also define a lower confidence bound,
\begin{align*}
\mathcal{H}_{t,k}^a(x)=\inf\left\lbrace \omega \in [0,1]: N_{t,k}^a(x) \cdot d\left(\hat{f}^a_{t,k}(x),\omega\right)\leq \theta \cdot \log t \right\rbrace - \varphi(t)\cdot r_{t,k}(x).
\end{align*}
Suppose that $\mathcal{H}_{t,k}^a(X_t)>f^a(X_t)$. Then since $t \geq \varphi^{-1}(\lambda)$ we have
\begin{align*}
N_{t,k}^a(x) \cdot d\left(\hat{f}^a_{t,k}(X_t),f^a(X_t)+\lambda \cdot r_{t,k}(X_t)\right)> \theta \cdot \log t.
\end{align*}
For $s \in [t-1]$ we define 
\begin{align*}
\epsilon_s &= \one\left\lbrace s\in \Gamma_{t,k}(X_t)\right\rbrace \cdot \one\left\lbrace \pi_s=a\right\rbrace\\
Z_s^{\mathcal{H}} & = \one\left\lbrace s\in \Gamma_{t,k}(X_t)\right\rbrace \cdot Y^a_s.
\end{align*}
Fix $\left\lbrace X_l\right\rbrace_{l \in [n]}$. By the Lipschitz property (Assumption \ref{lipschitzAssumption}) for all $s \in \Gamma_{t,k}(X_t)$ we have 
\begin{align*}
\E\left[Z_s^{\mathcal{H}}\hspace{1mm}\bigg|\left\lbrace X_l\right\rbrace_{l \in [n]}\right]&=f^a(X_s)\leq f^a(X_t)+ \lambda \cdot \rho\left(X_s,X_t\right).
\end{align*}
Hence, by Corollary \ref{klConcentrationInequality}, Appendix \ref{concentrationInequalitySec} we have
\begin{align*}
\Prob\left[ \mathcal{H}_{t,k}^a(X_t)>f^a(X_t) \hspace{1mm}\bigg|\left\lbrace X_l\right\rbrace_{l \in [n]}\right]
\end{align*}
\begin{align*}
&\leq \Prob\left[N_{t,k}^a(x) \cdot d\left(\hat{f}^a_{t,k}(X_t),f^a(X_t)+\lambda \cdot r_{t,k}(X_t)\right)> \theta \cdot \log t \hspace{1mm}\bigg|\left\lbrace X_l\right\rbrace_{l \in [n]}\right]\\
& \leq e \lceil \theta \cdot \log(t)^2\rceil t^{-\theta}.
\end{align*}
Similarly, by applying Corollary \ref{klConcentrationInequality} with $\left\lbrace \epsilon_s\right\rbrace_{s \in [t-1]}$ and $\left\lbrace Z_s^{\I}\right\rbrace_{s \in [t-1]}$ where $Z_s^{\I} = \one\left\lbrace s\in \Gamma_{t,k}(X_t)\right\rbrace \cdot (1-Y^a_s)$. We have,
\begin{align*}
\Prob\left[ \mathcal{I}_{t,k}^a(X_t)<f^a(X_t) \hspace{1mm}\bigg|\left\lbrace X_l\right\rbrace_{l \in [n]}\right]
\end{align*}
\begin{align*}
&\leq \Prob\left[N_{t,k}^a(x) \cdot d\left(\hat{f}^a_{t,k}(X_t),f^a(X_t)-\lambda \cdot r_{t,k}(X_t)\right)> \theta \cdot \log t \hspace{1mm}\bigg|\left\lbrace X_l\right\rbrace_{l \in [n]}\right]\\
& \leq e \lceil \theta \cdot \log(t)^2\rceil t^{-\theta}.
\end{align*}
Thus, by the total law of expectations we have
\begin{align*}
1-\Prob \left[\mathcal{H}_{t,k}^a(X_t)\leq f^a(X_t)\leq \I_{t,k}^a(X_t)\right] \leq 2e \lceil \theta \cdot \log(t)^2\rceil t^{-\theta}.
\end{align*}
By Pinsker's inequality, if $ N_{t,k}^a(x) \cdot d\left(\hat{f}^a_{t,k}(x),\omega\right)\leq \theta \cdot \log t$ then 
\begin{align*}
\big|\omega -\hat{f}^a_{t,k}(x) \big|\leq \sqrt{\frac{\theta \cdot \log t}{2\cdot N_{t,k}^a(x)}}.
\end{align*}
Hence, we have
\begin{align*}
\I_{t,k}^a(X_t)\leq \hat{f}^a_{t,k}(X_t)+\sqrt{\frac{\theta \cdot \log t}{2\cdot N_{t,k}^a(x)}}+\varphi(t) \cdot r_{t,k}(X_t)\leq \hat{f}^a_{t,k}(X_t)+U^a_{t,k}(X_t).
\end{align*}
Similarly, $\mathcal{H}_{t,k}^a(X_t)\geq \hat{f}^a_{t,k}(X_t)-U^a_{t,k}(X_t)$, so $\mathcal{H}_{t,k}^a(X_t)\geq \I_{t,k}^a(X_t)-2\cdot U^a_{t,k}(X_t)$.
Thus, for $t \geq \varphi^{-1}(\lambda)$, $\mathcal{H}_{t,k}^a(X_t)\leq f^a(X_t)\leq \I_{t,k}^a(X_t)$ implies $\mathcal{G}_{t,k}^a$, so
\begin{align*}
1-\Prob\left[\mathcal{G}_{t,k}^a\right] &\leq 1-\Prob \left[\mathcal{H}_{t,k}^a(X_t)\leq f^a(X_t)\leq \I_{t,k}^a(X_t)\right] \leq 2e \lceil \theta \cdot \log(t)^2\rceil t^{-\theta}.
\end{align*}

\end{proof}

\begin{proof}[Proof of  Theorem \ref{knnKlUCBRegretBound}] By Lemma \ref{applyConcentrationToKnnKlUCBLemma} we have
\begin{align*}
\sum_{t \in [n]}\left(1-\Prob\left[\mathcal{G}_t\right]\right)&\leq \varphi^{-1}(\lambda)+\sum_{a \in [A]}\sum_{t=\varphi^{-1}(\lambda)}^n\sum_{k \in [t-1]}\left(1-\Prob\left[\mathcal{G}_{t,k}^a\right]\right)\\
&\leq \varphi^{-1}(\lambda)+2e \cdot \sum_{a \in [A]}\sum_{t=\varphi^{-1}(\lambda)}^n\sum_{k \in [t-1]} \lceil \theta \cdot \left(\log t\right)^2 \rceil \cdot t^{-\theta}\\
&\leq \varphi^{-1}(\lambda)+2e \cdot A \cdot \sum_{t=1}^{\infty}\lceil \theta \cdot \left(\log t\right)^2 \rceil \cdot t^{1-\theta}.
\end{align*}
Given $\theta>2$ we have $\sum_{t=1}^{\infty}\lceil \theta \cdot \left(\log t\right)^2 \rceil \cdot t^{1-\theta}<\infty$. Hence, by applying Theorem \ref{generalisedIndexRegretBound}, the regret bound in Theorem \ref{knnKlUCBRegretBound} holds.
\end{proof}

%% file: Sections/ConcentrationInequality.tex
The following theorem is closely related to \cite[Theorem 11]{garivier2011kl}.

\begin{theorem}\label{legendreConcentrationInequality} Let $(Z_t)_{t \in [n]}$ be a sequence of real-valued random variables defined on a probability space $\left(\Omega,\mathcal{F},\Prob\right)$. Let $\left\lbrace \mathcal{F}_t\right\rbrace_{t \in \{0\}\cup[n]}$ be an increasing sequence of sigma fields such that for each $t$, $\sigma\left(Z_1,\cdots,Z_t\right)\subset \mathcal{F}_t$ and for $s>t$, $Z_s$ is independent from $\mathcal{F}_t$. Let $\left\lbrace \epsilon_t\right\rbrace_{t \in [n]}$ be a sequence of Bernoulli random variables such that $\sigma(\epsilon_t) \subset \mathcal{F}_{t-1}$. For each $t \in [n]$ we let 
\begin{align*}
S(t)=\sum_{s \in [t]}\epsilon_s \cdot Z_s,\hspace{1cm}N(t)=\sum_{s \in [t]}\epsilon_s,\hspace{1cm}\hat{\xi}_t=S(t)/N(t).
\end{align*}
Suppose we have a function $\phi:[0,\infty)\rightarrow \R$ with the following properties
\begin{itemize}
\item $\phi$ is twice differentiable with $\phi''(\rho)>0$ for all $\rho>0$,
\item For all $t \in [n]$ and $\rho \geq 0$ we have $\log\left(\E\left[\exp\left(\rho \cdot Z_t\right)\right]\right) \leq \phi(\rho)$,
\item For all $t \in [n]$ we have $\Prob\left[Z_t > \lim_{\rho \rightarrow \infty}\phi'(\rho)\right]=0$,
\item $\phi(0)=0$.
\end{itemize}
We define the Legendre transform $\phi^*:\R \rightarrow \R$ by $\phi^*(x):=\sup_{\rho \geq 0}\left\lbrace\rho \cdot x-\phi(\rho)\right\rbrace$. For all $\delta>0$ we have
\begin{align*}
\Prob\left[ N(n) \cdot \phi^*\left(\hat{\xi}_n\right)>\delta\right]\leq e \lceil \delta \log(n) \rceil \exp(-\delta).
\end{align*}
\end{theorem}
\begin{proof} Since $Z_{t+1}$ is independent from $\mathcal{F}_t$ and $\epsilon_{t+1}\in \{0,1\}$ is $\mathcal{F}_t$ measureable we have
\begin{align*}
\E\left[\exp\left(\rho \cdot \epsilon_{t+1} \cdot Z_{t+1}\right)|\mathcal{F}_t\right] &= \E\left[\left(1-\epsilon_{t+1}\right)+\epsilon_{t+1}\cdot \exp\left(\rho \cdot Z_{t+1}\right)|\mathcal{F}_t\right]\\
&=(1-\epsilon_{t+1})+\epsilon_{t+1} \cdot \E\left[\exp\left(\rho \cdot Z_{t+1}\right)\right]\\
& \leq (1-\epsilon_{t+1}) + \epsilon_{t+1} \cdot \exp(\phi(\rho)) = \exp\left( \epsilon_{t+1}\cdot  \phi(\rho)\right).
\end{align*}
For each $\rho \in \R$ we define $\left\lbrace W_t^{\rho}\right\rbrace_{t\in \{0\}\cup[n]}$ by $W_0^{\rho}=1$ and $W_t^{\rho}:=\exp\left(\rho \cdot S(t)-N(t)\cdot \phi(\rho)\right)$. Thus, $W_t^{\rho}$ is $\mathcal{F}_t$-measureable. Moreover, by the above we have
\begin{align*}
\E\left[W_{t+1}^{\rho}|\mathcal{F}_t\right] & = \E\left[\exp\left(\rho \cdot \epsilon_{t+1} \cdot Z_{t+1}-\epsilon_{t+1}\cdot \phi(\rho)\right)\cdot W_{t}^{\rho}|\mathcal{F}_t\right]\\
& = W_{t}^{\rho}\cdot \E\left[\exp\left(\rho \cdot \epsilon_{t+1} \cdot Z_{t+1}\right)|\mathcal{F}_t\right]\cdot \exp\left(-\epsilon_{t+1}\cdot \phi(\rho)\right)\leq W^{\rho}_t.
\end{align*}
Hence,  $\left\lbrace W_t^{\rho}\right\rbrace_{t\in \{0\}\cup[n]}$ is a super Martingale with respect to $\left\lbrace \mathcal{F}_t\right\rbrace_{t\in \{0\}\cup[n]}$.

By considering the derivative $\frac{\partial}{\partial \rho}\left(\rho \cdot x-\phi(\rho)\right) = x-\phi'(\rho)$ and noting that $\phi''>0$ on $\left(0,\infty\right)$ we see that for all $\rho \geq 0$ we have $\phi^*\left(\phi'(\rho)\right) = \rho \cdot \phi'(\rho)-\phi(\rho)$. In particular, $\phi^*\left(\phi'(0)\right) = 0$ since $\phi(0)=0$, and for all $\rho>0$, we have \[\frac{\partial}{\partial \rho}\left(\phi^*\left(\phi'(\rho)\right)\right)=\rho \cdot \phi''(\rho)>0.\]
Thus, $\lim_{\rho \rightarrow \infty}\phi^*\left(\phi'(\rho)\right)>0$.
Moreover, $(\phi^*) \circ (\phi'):[0,\infty)\rightarrow \left[0,\lim_{\rho\rightarrow \infty}\phi^*\left(\phi'(\rho)\right)\right)$ is an increasing bijection.

Now fix $\gamma>1$, to be determined later, and let \[\Delta:=\min\left\lbrace \lim_{\rho \rightarrow \infty}\phi^*\left(\phi'(\rho)\right),\delta\right\rbrace>0.\]
For each $q \in \{0\}\cup \N$ we let $t_q:=\lfloor (\delta/\Delta) \cdot \gamma^q \rfloor$ and let $Q:= \lceil \log(n)/\log \gamma \rceil$, so $t_Q\geq n$. Note also that if $N(n) \leq t_0 \leq \delta/\Delta \leq \delta \cdot \left(\lim_{\rho \rightarrow \infty}\phi^*\left(\phi'(\rho)\right)\right)^{-1}$ we have $\phi^*(\hat{\xi}_n)\leq \delta/N(n)$ with probability one, since $\Prob\left[Z_t > \lim_{\rho \rightarrow \infty}\phi'(\rho)\right]=0$ for each $t \in [n]$ and $\phi^*$ is everywhere non-decreasing.

Hence, if we let 
\begin{align*}
A_q:= \left\lbrace N(n) \cdot \phi^*\left(\hat{\xi}_n\right)>\delta\right\rbrace \cap \left\lbrace t_{q-1}< N(n) \leq t_q \right\rbrace,
\end{align*}
then we have
\begin{align*}
\Prob\left[ N(n) \cdot \phi^*\left(\hat{\xi}_n\right)>\delta \right] \leq \sum_{q=1}^Q \Prob\left[A_q \right].
\end{align*}
Now since $0<\Delta \leq  \lim_{\rho \rightarrow \infty}\phi^*\left(\phi'(\rho)\right)$ and $\gamma>1$, for each $q=1,\cdots,Q$ we may choose $\rho_q \in \left(0, \infty\right)$ so that $\phi^*(\phi'(\rho_q)) = \Delta \cdot \gamma^{-q}$. 

Hence, if $A_q$ holds then since $t_{q-1}< N(n) \leq t_q$ we must have
\begin{align*}
\phi^*(\phi'(\rho_q))= \frac{\Delta}{\gamma^q} \leq \frac{\delta}{N(n)} <\frac{\Delta}{\gamma^{q-1}} = \gamma \cdot \phi^*(\phi'(\rho_q)).
\end{align*}
Thus, as $N(n) \cdot \phi^*(\hat{\xi}_n)>\delta$ and $\phi^*$ is non-decreasing we have $\hat{\xi}_n>\phi'(\rho_q)$. Thus,
\begin{align*}
\rho_q \cdot S(n)-N(n)\cdot \phi(\rho_q)&= N(n) \cdot\left( \rho_q \cdot \hat{\xi}_n-\phi(\rho_q)\right)\\
& \geq N(n) \cdot\left( \rho_q \cdot \phi'(\rho_q)-\phi(\rho_q)\right)\\ &= N(n) \cdot \phi^*(\phi'(\rho_q)) > \frac{\delta}{\gamma}.
\end{align*}
Hence,
\begin{align*}
\Prob\left[A_q\right]\leq \Prob\left[W^{\rho_q}_n>\exp\left(\frac{\delta}{\gamma}\right)\right] \leq \E\left[W^{\rho_q}_n\right] \cdot \exp\left(-\frac{\delta}{\gamma}\right)\leq \exp\left(-\frac{\delta}{\gamma}\right), 
\end{align*}
by the super-Martingale property. Hence, for any $\gamma>1$ we have
\begin{align*}
\Prob\left[ N(n) \cdot \phi^*\left(\hat{\xi}_n\right)>\delta\right]\leq \bigg\lceil \frac{\log(n)}{\log \gamma} \bigg\rceil\cdot  \exp\left(-\frac{\delta}{\gamma}\right).
\end{align*}
Taking $\gamma = \delta/\left(\delta-1\right)$ completes the proof.
\end{proof}

\begin{corollary}\label{subGaussianConcentrationInequality} Let $(Z_t)_{t \in [n]}$ be a sequence of sub-Gaussian random variables, with $\E\left[\exp\left(\rho \cdot Z_t\right)\right]\leq \exp(\rho^2/2)$ for all $t \in [n]$ and $\rho \in \R$, defined on a probability space $\left(\Omega,\mathcal{F},\Prob\right)$. Let $\left\lbrace \mathcal{F}_t\right\rbrace_{t \in \{0\}\cup[n]}$ be an increasing sequence of sigma fields such that for each $t$, $\sigma\left(Z_1,\cdots,Z_t\right)\subset \mathcal{F}_t$ and for $s>t$, $Z_s$ is independent from $\mathcal{F}_t$. Let $\left\lbrace \epsilon_t\right\rbrace_{t \in [n]}$ be a sequence of Bernoulli random variables such that $\sigma(\epsilon_t) \subset \mathcal{F}_{t-1}$. For all $\delta>0$, with $N(n)$ and $\hat{\xi}_n$ as in the statement of Theorem \ref{legendreConcentrationInequality}, we have
\begin{align*}
\Prob\left[S(n) >\sqrt{2\delta\cdot N(n)}\right]\leq e \lceil \delta \log(n) \rceil \exp(-\delta).
\end{align*}
\end{corollary}
\begin{proof} Apply Theorem \ref{legendreConcentrationInequality} with $\phi(\rho)=\rho^2/2$.
\end{proof}

\begin{corollary}\label{klConcentrationInequality} Let $(Z_t)_{t \in [n]}$ be a sequence of random variables in bounded in $[0,1]$ defined on a probability space $\left(\Omega,\mathcal{F},\Prob\right)$ with $\max_{t \in [n]}\left\lbrace \E\left[ Z_t\right]\right\rbrace\leq \xi_{\max}$. Let $\left\lbrace \mathcal{F}_t\right\rbrace_{t \in \{0\}\cup[n]}$ be an increasing sequence of sigma fields such that for each $t$, $\sigma\left(Z_1,\cdots,Z_t\right)\subset \mathcal{F}_t$ and for $s>t$, $Z_s$ is independent from $\mathcal{F}_t$. Let $\left\lbrace \epsilon_t\right\rbrace_{t \in [n]}$ be a sequence of Bernoulli random variables such that $\sigma(\epsilon_t) \subset \mathcal{F}_{t-1}$. Let $N(n)$ and $\hat{\xi}_n$ be as in the statement of Theorem \ref{legendreConcentrationInequality}. For all $\delta>0$ we have
\begin{align*}
\Prob\left[ \hat{\xi}_n >\xi_{\max} \text{ \& }N(n) \cdot d\left(\hat{\xi}_n,\xi_{\max}\right)>\delta\right]\leq e \lceil \delta \log(n) \rceil \exp(-\delta).
\end{align*}
\end{corollary}
\begin{proof} 
Let $\phi(\rho):=\log\left(1+\xi_{\max}\cdot\left(\exp(\rho)-1\right)\right)$. Note that for any $z \in [\xi_{\max},1]$ we have $d(z,\xi_{\max})= \sup_{\rho \geq 0}\left\lbrace \rho \cdot z - \phi(\rho)\right\rbrace$. Since $\E\left[ Z_t\right]\leq \xi_{\max}$, for all $\rho \geq 0$ we have $\E\left[\exp\left(\rho \cdot Z_t\right)\right] \leq 1+\E\left[Z_t\right]\cdot\left(\exp(\rho)-1\right)\leq \exp(\phi(\rho))$. Hence, the corollary follows from Theorem \ref{legendreConcentrationInequality}.
\end{proof}

%% file: Sections/localisedRegretLemmas.tex
In this section we will prove lemmas \ref{breakRegretIntoSeparateArmsLemma} and \ref{regretAndCounts}.

\breakRegretIntoSeparateArmsLemma*
\begin{proof} We decompose the expected regret as follows
\begin{align*}
R_n(\pi)&=\sum_{t \in [n]}\left(Y^{\pi^*_{t}}_t-Y^{\pi_t}_t\right)\\
&\leq \sum_{a \in [A]}\left(\sum_{t\in[n]}\one\left\lbrace \mathcal{G}_t\right\rbrace \cdot \one\left\lbrace X_t \in \X \right\rbrace \cdot \one\left\lbrace \pi_t=a\right\rbrace \cdot\left(Y^{\pi^*_{t}}_t-Y^{\pi_t}_t\right)\right)\\&\hspace{1cm}+\sum_{t \in [n]}\left(1-\one\left\lbrace \mathcal{G}_t\right\rbrace\right)\cdot \left(Y^{\pi^*_{t}}_t-Y^{\pi_t}_t\right).
\end{align*}
Recall that $M:=\sup\left\lbrace \Delta^a(x): a \in [A],\hspace{2mm} x \in \X \right\rbrace$. Moreover, given any history $\D_{t-1}$, reward vector $X_t$ and arm $\pi_t$ we have,  
\begin{align*}
\E\left[Y^{\pi^*_{t}}_t-Y^{\pi_t}_t|\D_{t-1},X_t,\pi_t\right] &= f^*(X_t)-f^{\pi_t}(X_t)\leq M.
\end{align*}
Hence, $\E\left[Y^{\pi^*_{t}}_t-Y^{\pi_t}_t|\neg \mathcal{G}_t\right]\leq M$, since $\mathcal{G}_t$ is determined by $\mathcal{D}_{t-1}$, $X_t$, $\pi_t$, by the tower property. Thus, taking expectations in the above decomposition completes the proof of the lemma.
\end{proof}

\regretAndCounts*
\begin{proof}
By the definitions of $\tilde{R}_n^a(\pi,B)$ and $T^a_n(\pi,B)$ we have
\begin{align*}
\E&\left[\tilde{R}_n^a(\pi,B)\right]\\
&=\sum_{t \in [n]}\E\left[\E\left[\one\left\lbrace \mathcal{G}_t\right\rbrace \cdot \one\left\lbrace X_t \in B \right\rbrace \cdot \one\left\lbrace \pi_t=a\right\rbrace \cdot \left(Y^{\pi^*_{t}}_t-Y^{\pi_t}_t\right)| \mathcal{D}_{t-1},X_t,\pi_t\right]\right]\\
&=\sum_{t \in [n]}\E\left[\one\left\lbrace \mathcal{G}_t\right\rbrace \cdot \one\left\lbrace X_t \in B \right\rbrace \cdot \one\left\lbrace \pi_t=a\right\rbrace \cdot \E\left[Y^{\pi^*_{t}}_t-Y^{\pi_t}_t| \mathcal{D}_{t-1},X_t,\pi_t\right]\right]\\
& = \E\left[ \sum_{t \in [n]}\one\left\lbrace \mathcal{G}_t\right\rbrace \cdot \one\left\lbrace X_t \in B \right\rbrace \cdot \one\left\lbrace \pi_t=a\right\rbrace \cdot \left(f^*(X_t)-f^a(X_t)\right)\right]\\
& \leq \Delta^a(B) \cdot \E\left[\sum_{t \in [n]}\one\left\lbrace \mathcal{G}_t\right\rbrace \cdot \one\left\lbrace X_t \in B \right\rbrace \cdot \one\left\lbrace \pi_t=a\right\rbrace\right]\\
&= \Delta^a(B) \cdot \E\left[T^a_n(\pi,B)\right].
\end{align*}
\end{proof}

%% file: Sections/nestedPartitionsLemma.tex
In this section we prove Lemma \ref{nestedMetricCubesProp}. The proof utilises  ideas from (\cite{kaenmaki2012existence}).

\nestedMetricCubesProp*

\begin{myLemma}\label{epsSeparatedLemma} Suppose that $\epsilon>0$, $U\subseteq V \subseteq \X$ and $U$ is a maximal $\epsilon$-separated subset of $V$. Suppose further that there exists a function $g:V\rightarrow U$ such that for all $v \in V$, $\rho(v,g(v))=\min_{u \in U}\left\lbrace \rho(u,v)\right\rbrace$. Then for all $u \in U$, $\rho(u,g(u))<\epsilon$ and if $v \in V\backslash \left\lbrace g(u)\right\rbrace$ then $\rho(u,v)\geq \epsilon/2$.
\end{myLemma}
\begin{proof}
Follows from the definition of a maximal $\epsilon$-separated set.
\end{proof}

\begin{proof}[Proof of Lemma \ref{nestedMetricCubesProp}]
Let $\left\lbrace x_{q,i}\right\rbrace_{i\in [m_q]}$ be a finite maximal $(\delta/2)\cdot (1-r) \cdot r^q$ separated subset of $\X$. This is possible by  Assumption \ref{dimensionAssumption}. 
For $l \in [q-1]$ we let $\left\lbrace x_{l,i}\right\rbrace_{i\in [m_l]}$ be a maximal $(\delta/2)\cdot (1-r) \cdot r^l$ separated subset of $\left\lbrace x_{l+1,i}\right\rbrace_{i\in [m_{l+1}]}$, and define a function $g_l:\left\lbrace x_{l+1,i}\right\rbrace_{i\in [m_{l+1}]}\rightarrow \left\lbrace x_{l,i}\right\rbrace_{i\in [m_l]}$ by  $
g_l(x_{l+1,j}) = x_{l,i_j}$ where $i_j = \min\left\lbrace \text{argmin}_{i \in [m_l]}\left\lbrace \rho(z,x_{l,i_j})\right\rbrace\right\rbrace$.

The collection of sets $\left\lbrace Z_{l,i}:l \in [q],i \in [m_l]\right\rbrace$ is defined as follows. First define a partition $\left\lbrace Z_{q,i}\right\rbrace_{i \in [m_q]}$ by 
\begin{align*}
Z_{q,i}&:=B\left(x_{q,i},(\delta/2)\cdot (1-r) \cdot r^q\right)\backslash \bigcup_{j<i}Z_{q,j}.
\end{align*}
Then for $l=q-1,\cdots,1$ we define partitions $\left\lbrace Z_{l,i}\right\rbrace_{i \in [m_l]}$ by 
\begin{align*}
Z_{l,i}= \bigcup\left\lbrace Z_{l+1,j}:g_l(x_{l+1,j})=x_{l,i}\right\rbrace.
\end{align*}
Properties (1) and (2) in Proposition \ref{nestedMetricCubesProp} are immediate. 

We claim that for all $l \in [q]$ and $i\in[m_l]$, $Z_{l,i}\subseteq B\left(x_{l.i},(\delta/2)\cdot r^l\right)$. For $l=q$ this follows from the construction of $\left\lbrace Z_{q,i}\right\rbrace_{i \in [m_q]}$. For $l\in [q-1]$, we assume that the claim holds for $l+1$. Given $z \in Z_{q,i}$ for some $i\in [m_q]$, by construction we must have $z \in Z_{l+1,j}$ for some $j \in [m_{l+1}]$ with $g(x_{l+1,j})=x_{l,i}$. Hence, by assumption $\rho(z,x_{l+1,j})<(\delta/2)\cdot r^{l+1}$. Also, by Lemma \ref{epsSeparatedLemma} we have $\rho(x_{l+1,j},x_{l,i})\leq (\delta/2)\cdot (1-r)\cdot r^l$. Hence, $\rho(z,x_{l,i})<(\delta/2)\cdot r^l$, which proves the claim.

In addition we claim for all $l \in [q]$ and $i\in[m_l]$, \mbox{$B\left(x_{l,i},(\delta/4)\cdot (1-3r)\cdot r^l\right)\subseteq Z_{l,i}$}. Indeed, for $l=q$, it follows from the fact that $\left\lbrace x_{q,i}\right\rbrace_{i\in [m_q]}$ is $(\delta/2)\cdot (1-r) \cdot r^q$ separated that 
\begin{align*}
B\left(x_{q,i},(\delta/4)\cdot (1-3r)\cdot r^q\right)\subseteq B\left(x_{q,i},(\delta/4)\cdot r^q\right) \subseteq Z_{q,i}.
\end{align*}
For $l \in [q-1]$, $i \in [m_l]$ we consider $z \notin Z_{l,i}$. Take $j \in [m_{l+1}]$ so that $z \in Z_{l+1,j}$. Given the construction of $Z_{l,i}$ we have $g(x_{l+1,j})\neq x_{l,i}$. Hence, by Lemma \ref{epsSeparatedLemma} we have $\rho(x_{l+1,j}),x_{l,i}) \geq (\delta/4) \cdot (1-r)\cdot r^l$. Moreover, by the previous claim we have $\rho(z,x_{l+1,j})<(\delta/2)\cdot r^{l+1}$. Hence, we have $\rho(z,x_{l,i})\geq (\delta/4)\cdot (1-3r)\cdot r^l$. which completes the proof of the second claim.

Now take $l \in [q]$, $i \in [m_l]$. Since $Z_{l,i}\subseteq B\left(x_{l.i},(\delta/2)\cdot r^l\right)$ we must have $\diam(Z_{l,i})\leq \delta \cdot r^l$. By Assumption \ref{dimensionAssumption}, combined with the fact that  \mbox{$B\left(x_{l,i},(\delta/4)\cdot (1-3r)\cdot r^l\right)\subseteq Z_{l,i}$} gives $\mu\left(Z_{l,i}\right) \geq C_{\Xdim}\cdot \left( (\delta/4)\cdot (1-3r)\cdot r^l\right)^d$. Hence, property (3) in  Proposition \ref{nestedMetricCubesProp} also holds.

\end{proof}

%% file: Sections/experimentalDetails.tex
In this section we give a detailed account of the experimental procedure for the empirical results in Section \ref{experimentalSec}. We constructed a synthetic $A$-armed bandit problem with covariates on a $d$-dimensional sub-manifold within a $D$-dimensional feature space $\mathbb{R}^D$ as follows: 

To construct a marginal distributions $\mu$ on $\mathbb{R}^D$, supported on a $d$-dimensional sub-manifold $\mathcal{M}\subset \mathbb{R}^D$, we first construct an affine map $\phi:\R^d\rightarrow \R^D$. We do this by randomly generating $d$ orthonormal $D$-dimensional vectors $\{\bm{u}_1,\cdots,\bm{u}_d\}\subset \R^D$, letting
$\tilde{\phi}(z)=\sum_{l=1}^dz_l\cdot \bm{u}_l$ for $z = \left(z_l\right)_{l=1}^d \in \mathbb{R}^d$, and letting $\phi(z) = \tau \circ \tilde{\phi}$, where $\tau:\mathbb{R}^D \rightarrow \mathbb{R}^D$ is a similarity mapping such that ${\phi}\left([0,1]^d\right)\subseteq [0,1]^D$. It follows that $\mathcal{M}:={\phi}\left([0,1]^d\right)\subseteq [0,1]^D$ is a compact subset of the $d$-dimensional manifold $\phi(\R^d)$ and the inverse $\phi^{-1}:\mathcal{M}\rightarrow [0,1]^d$ is well-defined. We construct a measure $\tilde{\mu}$ on $[0,1]^d$ by taking \[S:=\bigcup_{\vec{\omega}\in [5]^d}\left\lbrace z \in \R^d: \|10 \cdot z-2\vec{\omega}+\bm{1}_d\|_{\infty}\leq 1/5\right\rbrace,\] and letting $\tilde{\mu}$ be the uniform measure on $S$. We then obtain $\mu$ supported on $\mathcal{M}$ by $\mu:= \tilde{\mu} \circ \phi^{-1}$. The use of $S$ in the construction of $\mu$ ensures that that we have large margins $\Delta(x)$ with high probability. We construct reward functions $f^a$ for each arm $a \in [A]$ as follows: First we define a kernel function $h:\mathbb{R}^d\rightarrow \mathbb{R}$ by $h(x):= \max\left\lbrace 1-\|x\|_{\infty},0\right\rbrace$. For each arm $a \in [A]$, and each vector $\vec{\omega} \in [5]^d$ we select $\zeta^a({\vec{\omega}})\in \{-1,+1\}$ randomly (i.i.d with probability 0.5). We construct a Lipschitz function $\tilde{f}^a:[0,1]^d\rightarrow [0,1]$ by 
\begin{align*}
\tilde{f}^a(z):= \frac{1}{2}+\frac{1}{10}\cdot \sum_{\omega \in [5]^d} \zeta^a({\vec{\omega}}) \cdot h\left(10z-2\vec{\omega}+\bm{1}_d\right),
\end{align*}
where $\bm{1}_d$ is a $d$-dimensional vector consisting entirely of ones. Finally, we define \mbox{$f^a:\mathcal{M} \rightarrow [0,1]$} by $f^a(x)=\tilde{f}^a\left(\phi^{-1}(x)\right)$. We generate $(X,Y^1,\cdots,Y^A) \in \mathcal{M}\times \{0,1\}^A$ by $X \sim \mu$ and $\E\left[Y^a|X=x\right]=f^a(X)$. 

In our experiments we consider four bandit scenarios and four algorithms. In each bandit scenario the dimension of the manifold $d=2$ and the number of arms $A=2$. We vary the dimension of the ambient feature space $D$ over the scenarios by taking $D \in \{2,5,10,15\}$. We compare four algorithms: The UCBogram \cite{rigollet2010nonparametric}, the ABSE algorithm \cite{perchet2013multi}, the K-NN UCB algorithm with $\theta=2$, $\varphi \equiv 1$ and the K-NN KL-UCB algorithm with $\theta=1$, $\varphi \equiv 1$. For each of the sixteen combinations of bandit scenario and algorithm we conduct fifty runs, with fifty different random seeds, each time with a horizon of one hundred thousand.

%% file: Sections/exampleDiadicMeasureBadlyBehaved.tex
The Adaptively Binned Successive Elimination requires the following assumption.

\begin{assumption}[Dyadic cubes assumption]\label{dyadiccubesAss} There exists $C_d,d>0$ such that for every $B\subset [0,1]^D$ of the form $B=2^{-q}\cdot \prod_{i=1}^D[z_i,z_i+1]$ with $z_1,\cdots,z_D,q\in \N\cup\{0\}$, we have either $\mu(B)\geq C_d\cdot \diam(B)^d$ or $\mu(B)=0$.
\end{assumption}

In the following example Assumption \ref{dimensionAssumption} holds yet Assumption \ref{dyadiccubesAss} does not.
\begin{example} We define $\theta = \sum_{n = 1}^{\infty}2^{-n!}$, take $I_{\theta}=[0,\theta]$ and let $\mu_{\theta}$ denote the normalised Lebesgue measure on $I_{\theta}$. For any $x\in \supp(\mu_{\theta})=I_{\theta}$, and $r\in (0,\theta)$, $B(x,r)\cap I_{\theta}$ is an interval of diameter at least $r$, so we have $\mu_{\theta}(B(x,r))\geq r/\theta$. Hence, Assumption \ref{dimensionAssumption} holds with $C_d=\theta^{-1}$ and $d=1$. On the other hand, Assumption \ref{dyadiccubesAss} does not hold. Indeed given $q\in \N$, we consider the dyadic interval $B_q:=[\sum_{n=1}^q2^{-n!},\sum_{n=1}^q2^{-n!}+2^{q!}]$. Then $\diam(B_q)=2^{-q!}$. However, $B_q\cap I_{\theta}=[\sum_{n=1}^q2^{-n!},\sum_{n=1}^{\infty}2^{-n!}]$, so
\begin{align*}
\mu(B_q) = \theta^{-1}\cdot \diam(B_q) = \theta^{-1}\cdot \sum_{n=q+1}2^{-n!} \leq (2/\theta) \cdot 2^{-(q+1)!} = (2/\theta) \cdot \diam(B_q)^{q+1}.
\end{align*}
Consequently, given any $C_d,d>0$ we can take $q\in \N$ sufficiently large that $q>d$ and $(2/\theta)\cdot 2^{-q!}<C_d$. It follows that whilst $\mu(B_q)\neq \emptyset$ we do have $\mu(B_q) \leq (2/\theta) \cdot 2^{-q!}\cdot \diam(B_q)^{q+1}<C_{d}\cdot \diam(B_q)^d$.
\end{example}

%% file: Sections/manifoldsAndTheLowerBound.tex
In this section we shall recall some results regarding manifolds. This will serve two proposes. Firstly, we will make precise the sense in which Assumption \ref{dimensionAssumption} holds for all well-behaved measures $\mu$ supported on a $d$-dimensional submanifold of Euclidean space. Secondly, we will demonstrate that the regret bounds in Theorems \ref{knnUCBRegretBound} and \ref{knnKlUCBRegretBound} are minimax optimal up to logarithmic factors.

\subsection{Manifolds, reach and regular measures}

Suppose we have a $C^{\infty}$-smooth sub-manifold of $\mathcal{M} \subset \mathbb{R}^D$ of dimension $d$ (see \cite{lee2006riemannian}). We shall make use of the concept of reach $\tau$ introduced by \cite{federer1959curvature} and investigated by  \cite{niyogi2008finding}. The reach $\tau$ of a manifold $\mathcal{M}$ is defined by
\begin{align*}
\tau:= \sup\left\lbrace r>0:\forall z\in \R^D \text{  }\inf_{q\in \mathcal{M}}\left\lbrace \|z-q\|_2\right\rbrace <r \implies 
\exists\text{! } p \in \mathcal{M},\hspace{2mm}\|z-p\|_2=
\inf_{q\in \mathcal{M}}\left\lbrace \|z-q\|_2\right\rbrace
\right\rbrace. \end{align*}
Note that \cite{niyogi2008finding} refers to the condition number $1/\tau$, which is the reciprocal of the reach $\tau$. We let $V_{\mathcal{M}}$ denote the Riemannian volume.

\begin{defn}[Regular sets and measures]\label{regSetsAndMeasuresDef}
Suppose we have a measure $\upsilon$ on the metric space $(\X,\rho)$. A subset $A\subset \X$ is said to be a $(c_0,r_0)$-regular set with respect to the measure $\upsilon$ if for all $x\in A$ and all $r\in (0,r_0)$ we have $\upsilon\left(A\cap B_r(x)\right)\geq c_0\cdot \upsilon\left(B_r(x)\right)$, where $B_r(x)$ denotes the open metric ball of  radius $r$, centred at $x$. A measure $\mu$ with support $\supp(\mu)\subset \X$ is said to be $(c_0,r_0,\nu_{\min},\nu_{\max})$-regular measure with respect to $\upsilon$ if $\supp(\mu)$ is a $(c_0,r_0)$-regular set with respect to $\upsilon$ and $\mu$ is absolutely continuous with respect to $\upsilon$ with Radon-Nikodym derivative $\nu(x)=d\mu(x)/d\upsilon(x)$, such that for all $x\in \supp(\mu)$ we have $\nu_{\min}\leq \nu(x)\leq \nu_{\max}$.  
\end{defn}

\subsection{The dimension assumption on manifolds}

In this section we justify Assumption \ref{dimensionAssumption} showing that it holds whenever the marginal $\mu$ is regular with respect to a $d$-dimensional manifold. The proof follows straightforwardly from \cite{eftekhari2015new}.

\begin{prop}\label{manifoldImpliesDimensionAssumption} Let $\mathcal{M}\subseteq \R^D$ be a $C^{\infty}$-smooth compact sub-manifold of dimension $d$ and reach $\tau$. Suppose that $\mu$ is a $(c_0,r_0,\nu_{\min},\nu_{\max})$-regular measure with respect to $V_{\mathcal{M}}$. Then $\mu$ satisfies the dimension assumption (Assumption \ref{dimensionAssumption}) with constants $R_{\mathcal{X}}=\min\left\lbrace \tau/4,r_0\right\rbrace$, $d$ and $C_d= \nu_{\min}\cdot c_0 \cdot v_d \cdot 2^{-d}$, where $v_d$ denotes the Lebesgue measure of the unit ball in $\R^d$.
\end{prop}

\begin{proof}
Take $x \in \supp(\mu)$ \& $r \in \left(0,R_{\mathcal{X}}\right)$. By \cite[Lemma 12]{eftekhari2015new} we have
\begin{align}\label{lowerBoundFromEftekhariWakin}
V_{\mathcal{M}}\left(B_r(x)\right) \geq \left(1-\frac{r^2}{4\tau^2}\right)^{\frac{d}{2}}\cdot v_{d}\cdot r^{d} \geq v_{d}\cdot 2^{-d}\cdot r^{d}.
\end{align}
Moreover, since $\mu$ is $(c_0,r_0,\nu_{\min},\nu_{\max})$-regular we have
\begin{align*}
\mu\left(B_r(x)\right) & \geq \nu_{\min}\cdot V_{\mathcal{M}}\left(B_r(x) \cap \supp(\mu)\right)\\
& \geq \nu_{\min}\cdot c_0 \cdot V_{\mathcal{M}}\left(B_r(x)\right).
\end{align*}
Combining with (\ref{lowerBoundFromEftekhariWakin}) proves the proposition.
\end{proof}

\subsection{A lower bound on regret for bandits on manifolds}

The following result demonstrates that the regret bounds in Theorems \ref{knnUCBRegretBound} and \ref{knnKlUCBRegretBound} are minimax optimal up to logarithmic factors, for all sufficiently well-behaved manifolds. The theorem follows straightforwardly from the proof of \cite[Proposition A.1]{pmlr-v76-reeve17a}, which generalises \cite[Theorem 3.5]{audibert2007fast} to embedded manifolds.

\begin{theorem}\label{lowerBound}
Let $\mathcal{M} \subset \R^D$ be a compact $C^{\infty}$ sub-manifold of dimension $d$ and reach $\tau$, and take $A \geq 2$. There exists a universal positive constant $Z>0$ and positive constants $C_0,R_0,V_-,V_+>0$ determined by $\gamma,\tau$ such that for all $c_0 \in \left(0,C_0\right)$, $r_0 \in \left(0,R_0\right)$, $\delta_{\alpha} \in \left(0,Z\right)$, $\alpha \in \left(0,d\right)$, $C_{\alpha}>0$, $\lambda>0$, there exists a constant $C>0$, depending solely upon $(d,\tau)$, $(c_0,r_0,\nu_{\min},\nu_{\max})$, $\left(\alpha,\delta_{\alpha},C_{\alpha}\right)$ and $\lambda$ such that the following holds: Given any policy $\pi$ and $n \in \N$ there exists a distribution $\Prob$ on pairs $(X,Y)$ with $X \in \R^D$ and $Y=(Y^a)_{a \in [A]} \in \{0,1\}^A$ such that the marginal over $X$, $\mu$ is $(c_0,r_0,\nu_{\min},\nu_{\max})$-regular with respect to $V_{\mathcal{M}}$ (so $\Prob$ satisfies Assumption \ref{dimensionAssumption} by Lemma \ref{manifoldImpliesDimensionAssumption}), the reward functions $f^a$ are $\lambda$-Lipschitz  (Assumption \ref{lipschitzAssumption}) and $\Prob$ satisfies the margin condition with constants $\delta_{\alpha},C_{\alpha},\alpha$ (Assumption \ref{marginAssumption}) and
\begin{align*}
\E\left[R_n(\pi)\right] \geq C \cdot  n^{1-\frac{\alpha+1}{d+1}}.
\end{align*}
\end{theorem}

\begin{proof}
Let $\mathcal{P}$ denote the set of all distributions $\Prob$ on $(X,Y)$ with $X \in \R^D$ and $Y=(Y^a)_{a \in [A]} \in \mathcal{Y}:= \{0,1\}^A$ such that:
\begin{enumerate}
    \item The marginal of $\Prob$ over $X$ is $(c_0,r_0,\nu_{\min},\nu_{\max})$-regular with respect to $V_{\mathcal{M}}$,
    \item The reward functions $f^a:x \mapsto \E\left[Y^a|X=x\right]$ are $\lambda$-Lipschitz,
    \item $\Prob$ satisfies the margin condition with constants $\delta_{\alpha},C_{\alpha},\alpha$.
\end{enumerate}
Let $\mathcal{Y}_{\text{class}}:= \left\lbrace Y=(Y^a)_{a \in [A]} \in \mathcal{Y}: \sum_{a \in [A]}Y^a = 1\right\rbrace$ and let $\mathcal{P}_{\text{class}}:=\left\lbrace \Prob \in  \mathcal{P}: \Prob\left[Y \in \mathcal{Y}_{\text{class}}\right]=1\right\rbrace$. Let $\phi$ be a supervised classification procedure. We may view $\phi$ as a map from pairs $\left(\mathcal{F}_{n-1},X_n\right)$, consisting of data set $\mathcal{F}_{n-1}=\left\lbrace \left(X_t,Y_t\right)
\right\rbrace_{t\in [n-1]} \in \left(\mathcal{M}\times \mathcal{Y}_{\text{class}} \right)^{n-1}$ and a covariate $X_n \in \mathcal{M}$, to an output $\phi\left(\mathcal{F}_{n-1},X_n\right) \in [A]$. From the proof of \cite[Proposition A.1]{pmlr-v76-reeve17a} we see that there exists a constant $C>0$, depending solely upon $(d,\tau)$, $(c_0,r_0,\nu_{\min},\nu_{\max})$, $\left(\alpha,\delta_{\alpha},C_{\alpha}\right)$ and $\lambda$ together with a finitely supported probability measure $p$ on $\mathcal{P}_{\text{class}}$ such that for all classification procedures $\phi$ we have
\begin{align*}
&\int_{\mathcal{P}_{\text{class}}}\left(\int_{\left(\mathcal{M}\times \mathcal{Y}_{\text{class}} \right)^{n}}\left(Y_n^{\pi_*\left(X_n\right)}-Y_n^{\phi\left(\mathcal{F}_{n-1},X_n\right)}\right) d\Prob^n\left(\mathcal{F}_n\right)\right)dp\left(\Prob\right)
\\&=
\int_{\mathcal{P}_{\text{class}}}\left(\int_{\left(\mathcal{M}\times \mathcal{Y}_{\text{class}} \right)^{n}}\mathds{1}\left\lbrace Y_n^{\phi\left(\mathcal{F}_{n-1},X_n\right)}\neq 1\right\rbrace-\mathds{1}\left\lbrace Y_n^{\pi_*\left(X_n\right)}\neq 1\right\rbrace d\Prob^{n}\left(\mathcal{F}_{n}\right)\right)dp\left(\Prob\right)\\
&\geq C \cdot (n-1)^{-\frac{\alpha+1}{d+2}} \geq  C \cdot n^{-\frac{1+\alpha}{2+d}}.
\end{align*}
Now let $\pi$ be any bandit policy. For each $t \in [n]$ we may convert $\pi$ into a classification procedure $\phi_t:\left(\mathcal{F}_{n-1},X_n\right)\mapsto \phi_t\left(\mathcal{F}_{n-1},X_n\right) \in [A]$ by first applying $\pi$ to $\left\lbrace \left(X_s,Y_s\right)\right\rbrace_{s \in [t-1]} \subset \mathcal{F}_{n-1}$ and letting $\phi_t\left(\mathcal{F}_{n-1},X_n\right) = \pi_t\left(X_n\right)$. Hence, for each $t \in [n]$ we have,
\begin{align*}
\int_{\mathcal{P}_{\text{class}}}\left(\int_{\left(\mathcal{M}\times \mathcal{Y}_{\text{class}} \right)^{n}}\left(Y_n^{\pi_*\left(X_n\right)}-Y_n^{\pi_t(X_n)}\right) d\Prob^n\left(\mathcal{F}_n\right)\right)dp\left(\Prob\right)
 \geq  C \cdot n^{-\frac{1+\alpha}{2+d}}.
\end{align*}
By symmetry, for each $t\in [n]$ we have,
\begin{align*}
\int_{\mathcal{P}_{\text{class}}}\left(\int_{\left(\mathcal{M}\times \mathcal{Y}_{\text{class}} \right)^{n}}\left(Y_t^{\pi_*\left(X_t\right)}-Y_t^{\pi_t(X_t)}\right) d\Prob^n\left(\mathcal{F}_n\right)\right)dp\left(\Prob\right)
 \geq  C \cdot n^{-\frac{1+\alpha}{2+d}}.
\end{align*}
Hence, by Fubini's theorem
\begin{align*}
\int_{\mathcal{P}_{\text{class}}} \E\left[ R_n(\pi)\right] dp\left(\Prob\right) &= \int_{\mathcal{P}_{\text{class}}}\left(\int_{\left(\mathcal{M}\times \mathcal{Y}_{\text{class}} \right)^{n}}\sum_{t \in [n]}\left(Y_t^{\pi_*\left(X_t\right)}-Y_t^{\pi_t(X_t)}\right) d\Prob^n\left(\mathcal{F}_n\right)\right)dp\left(\Prob\right)\\
& =  \sum_{t \in [n]}\int_{\mathcal{P}_{\text{class}}}\left(\int_{\left(\mathcal{M}\times \mathcal{Y}_{\text{class}} \right)^{n}}\left(Y_t^{\pi_*\left(X_t\right)}-Y_t^{\pi_t(X_t)}\right) d\Prob^n\left(\mathcal{F}_n\right)\right)dp\left(\Prob\right)\\
& \geq C \cdot n^{1-\frac{\alpha+1
}{d+2}}.
\end{align*}
In particular, there must exist some $\Prob \in \mathcal{P}_{\text{class}} \subset \mathcal{P}$ with $ \E\left[ R_n(\pi)\right] \geq C \cdot n^{1-\frac{\alpha+1
}{d+2}}$.
\end{proof}